\pdfoutput=1

\documentclass[11pt]{article}

\usepackage[final]{acl}

\usepackage{times}
\usepackage{latexsym}

\usepackage[T1]{fontenc}

\usepackage[utf8]{inputenc}

\usepackage{microtype}

\usepackage{inconsolata}

\usepackage{graphicx}

\usepackage{amsmath}
\usepackage{arydshln} 
\usepackage{booktabs}
\usepackage{amsmath}
\usepackage{bm}
\usepackage{tikz}
\usepackage{multirow}
\usepackage{colortbl}
\usepackage{wrapfig}
\usepackage{amssymb}
\usepackage{pifont}
\usepackage{xcolor} 
\usepackage{tabularray}
\usepackage{tabularx}
\usepackage{booktabs}
\usepackage{hyperref}
\usepackage{lipsum}

\definecolor{codeblue}{rgb}{0.25,0.5,0.5}
\definecolor{myblue}{rgb}{0.88,0.98,1}
\definecolor{mygreen}{rgb}{0.92, 1.0, 0.92}
\definecolor{myred}{rgb}{1, 0.9, 0.9}
\definecolor{mygray}{gray}{0.95}
\newcommand{\colorrect}[1]{\textcolor{#1}{\ding{110}}}
\definecolor{Highlight}{HTML}{E8F8F5}
\definecolor{midgreen}{HTML}{589d62}
\definecolor{midblue}{HTML}{69a3f1}
\definecolor{darkgreen}{HTML}{146038}
\definecolor{darkblue}{HTML}{143b59}
\definecolor{hotpink}{RGB}{59, 115, 227}

\title{Scalable Vision Language Model Training via High Quality Data Curation}

\author{
    Hongyuan Dong\textsuperscript{*}, Zijian Kang\textsuperscript{*}, Weijie Yin\textsuperscript{*}, Xiao Liang, Chao Feng\textsuperscript{\dag}, Jiao Ran\\
    ByteDance Douyin Content Group \\
    \texttt{\{donghongyuan.dousia, zijian.kang, yinweijie\}} \\
    \texttt{\{liangxiao.ilx, chaofeng.zz, ranjiao\}@bytedance.com}
}

\begin{document}

\maketitle
\newcommand\blfootnote[1]{%
    \begingroup
    \renewcommand\thefootnote{}\footnote{#1}%
    \addtocounter{footnote}{-1}%
    \endgroup
}
\blfootnote{* Equal contribution.}
\blfootnote{\dag Email corresponding}

\begin{abstract}
In this paper, we introduce \textit{\textbf{SAIL}-VL} (\textit{\textbf{S}c\textbf{A}lable Vision Language Model Tra\textbf{I}ning via High Qua\textbf{L}ity Data Curation}), an open-source vision language model (VLM) series achieving state-of-the-art (SOTA) performance in 2B and 8B parameters.  
The following three key improvements contribute to SAIL-VL's leading performance:
(1) Scalable high-quality visual understanding data construction: 
We implement a data construction pipeline to enable hundred-million-scale high-quality recaption data annotation.
The resulted dataset SAIL-Caption is validated to be of the highest data quality compared with opensource datasets. 
(2) Scalable Pretraining with High-Quality Visual Understanding Data: 
We scale SAIL-VL's pretraining budget up to 655B tokens and show that even a 2B VLM benefits from scaled up training data sizes, exhibiting logarithmic data size scaling laws in benchmark performance. 
(3) Scalable SFT via data quantity and complexity scaling: 
We curate a high-quality SFT dataset collection with leading data quantity scaling effectiveness and demonstrate that training with progressively higher-complexity data surpasses baseline one-stage training by a large margin. 

SAIL-VL series models achieve the highest average score in 18 widely used VLM benchmarks in our evaluation, with the 2B model takes the top position over VLMs of comparable sizes on OpenCompass 2024 (\href{https://rank.opencompass.org.cn/leaderboard-multimodal}{https://rank.opencompass.org.cn/leaderboard-multimodal}), demonstrating robust visual comprehension abilities. 
SAIL-VL series models are released at HuggingFace (\href{https://huggingface.co/BytedanceDouyinContent}{https://} \href{https://huggingface.co/BytedanceDouyinContent}{huggingface.co/BytedanceDouyinContent}). 

\end{abstract}

\begin{figure*}[t]
\centering
\resizebox{\textwidth}{60mm}{
\begin{tikzpicture}
\draw (0,0) node[inner sep=0] {\includegraphics[width=\columnwidth, trim={0cm 4.8cm 1cm 0cm}, clip]{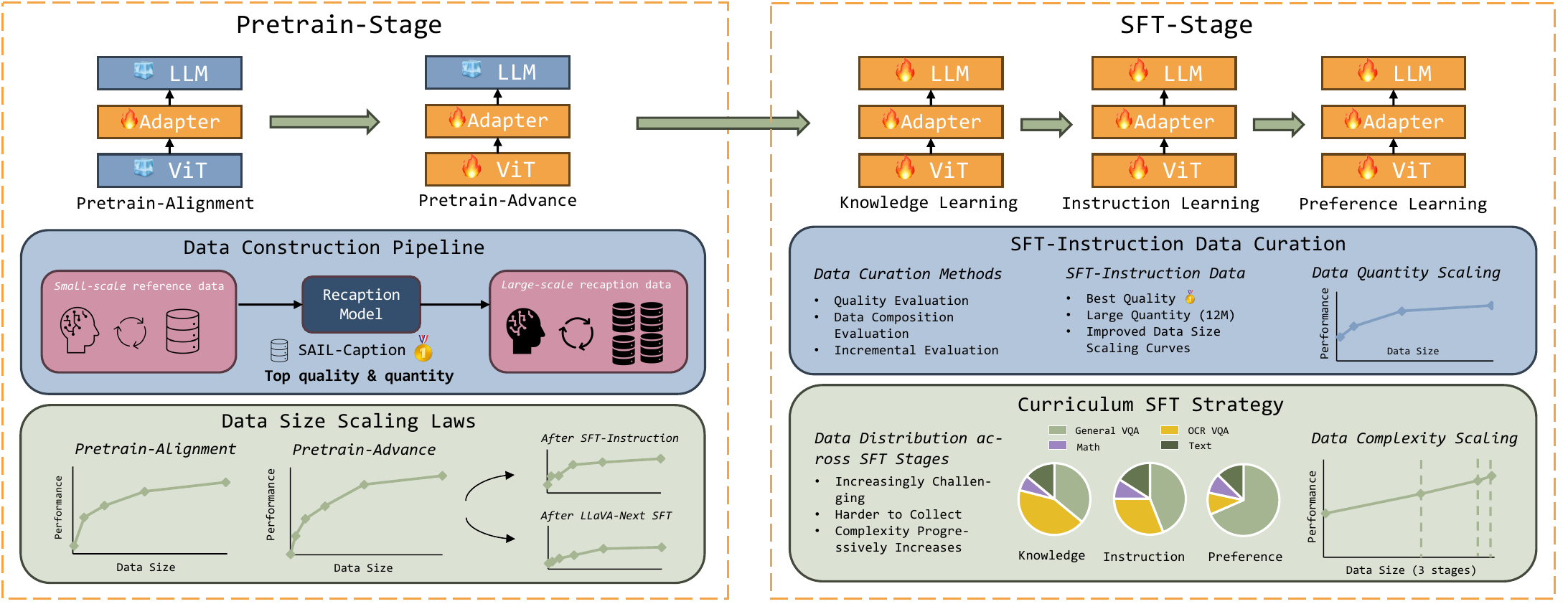}};
\end{tikzpicture}
}
\caption{
SAIL-VL's overall data construction and model training pipeline, as well as data size scaling laws observed in our large-scale VLM training experiments.
}
\label{fig: pipeline}
\vspace{-2mm}
\end{figure*}

\section{Introduction}
Researches in large vision language models (VLMs)~\cite{liu2024llavanext,li2024llava,yao2024minicpm,wang2024qwen2,gu2024infinity,chen2024far,chen2024expanding} have made significant progress in recent years, facilitating various vision tasks via language interactions. 
Due to the memory and computational constraints in model deployment, training compact VLMs with robust visual comprehension performance has become a popular research field recently~\cite{smolvlm,chen2023pali,yao2024minicpm,li2024mini,gao2024mini}.
However, how to make optimal use of publicly available resources to unlock the potential of compact VLMs remains an unanswered question. 
We attribute the suboptimal performance of recent lightweight vision language models to their limited fundamental visual understanding abilities and unsatisfactory instruction following performance. 


The fundamental visual understanding abilities of VLMs are typically established via large-scale pretraining, which necessitates not only substantial training budgets, but also a sufficient amount of high-quality visual understanding data to take effect.
Recently proposed VLMs, such as LLaVA series~\cite{liu2024llavanext,li2024llava,chen2024allava}, conduct light-weight pretraining with a limited amount of low-quality caption data, and therefore suffer from suboptimal visual understanding abilities which hinder subsequent visual instruction tuning.
MiniCPM-V-2.5~\cite{yao2024minicpm} and Qwen2-VL~\cite{wang2024qwen2} allocate hundreds of billions of tokens' computation budgets to the pretraining stage, but the limited visual understanding data quality undermines their visual understanding performance. 
More importantly, despite the large amount of resources consumed in pretraining, existing works do not provide reliable conclusions to understand how pretraining budgets and data quality influence VLM performance.


During the supervised fine-tuning (SFT) stage, VLM's visual understanding capabilities are generalized to instruction following tasks. 
However, how to make optimal use of high-quality visual instruction tuning datasets remains unexplored.
To obtain SFT data collections with higher quality, recent works focus on adjusting the data distribution across various domains and formats~\cite{li2024llava,chen2024far,chen2024expanding,yao2024minicpm}. 
Infinity-MM~\cite{gu2024infinity} further explores enhancing the data efficiency of visual instruction tuning datasets with a multi-stage SFT strategy, obtaining promising performance scaling results. 
Despite the promising results of these works, there still lacks widely acknowledged methodologies to determine the distribution of SFT dataset collections or allocation of SFT stages.



To address the above issues, we propose SAIL-VL, an opensource vision language model series in 2B and 8B parameters with state-of-the-art (SOTA) performance.
SAIL-VL is trained through several pretraining and SFT stages. 
We first establish SAIL-VL's basic visual understanding abilities via large-scale pretraining. 
To explore how pretraining computation budgets and data quality influence VLM performance, we scale up VLM pretraining to 655B tokens with \textit{SAIL-Caption}, our synthesized large-scale detail caption dataset with top data quality compared to opensource alternatives. 
During the following SFT stages, we train SAIL-VL on our customized SFT data collection which outperforms opensource datasets markedly in data quality. 
SAIL-VL is trained in a curriculum learning paradigm of three stages, leading to improved data efficiency and model performance. 
The resulting SAIL-VL-2B and 8B models achieve new SOTA performance in 18 widely used VLM benchmarks. 

We summarize the key contribution of this research as below:


\textbf{(1)} We implement a data construction pipeline for scalable high-quality visual understanding data construction, equipped with which we construct SAIL-Caption, which is of large quantity and the highest quality compared with opensource datasets. 

\textbf{(2)} We scale up SAIL-VL's pretraining data size to 655B tokens, and report logarithmic model performance scaling laws w.r.t. training data sizes.
To the best of our knowledge, this is the first time that data size scaling laws for VLM pretraining are proposed and discussed.

\textbf{(3)} We elaborate on the methodologies for high-quality SFT data curation, and demonstrate the effectiveness of the curriculum SFT strategy. 
Our SAIL-VL series models achieve top-ranked performance in our evaluation on 18 opensource VLM benchmarks. 


\section{Model Training Pipeline}
In this section, we introduce SAIL-VL's training strategy as shown in Fig~\ref{fig: pipeline}.
Starting from InternViT~\cite{chen2023internvl} and Qwen-2.5~\cite{qwen2.5} series models, SAIL-VL is pretrained for visual understanding and adapted to instruction following tasks in a total of five training stages.

\begin{table*}[t]
\small
\resizebox{\textwidth}{15mm}{
\begin{tabular}{l c c c c c c c c c}
\toprule
Dataset & Language & \# Sample & Avg. Len. & Quality & Uni. 2-gram & Uni. 3-gram & Uni. Noun & Uni. Verb & Uni. Adj.
\\
\midrule

$\text{SAIL-Caption}_{\text{EN}}$ & EN & 225,000,000 & 87.86 & - & 14.34 & 33.40 & 0.6980 & 0.0722 & 0.4719 \\
$\text{SAIL-Caption}_{\text{CN}}$ & CN & 75,000,000 & 156.95 & - & 11.71 & 32.10 & 1.032 & 0.3238 & 0.0625 \\

\midrule

DataComp-LLaVA-Caption~\citeyearpar{li2024if} & EN & 940,891,257 & 48.37 & 70.0 & 8.400 & 18.23 & 0.4728 & 0.0333 & 0.2801 \\
\rowcolor{gray!10}
$\text{SAIL-Caption-DataComp}_{\text{Subset}}$ & EN & 10,000 & 83.08 & 87.2 & 13.20 & 30.49 & 0.6354 & 0.0616 & 0.4627 \\

SA1B-QwenVL-Caption~\citeyearpar{sa1bqwen} & CN & 8,631,495 & 130.3 & 74.6 & 7.450 & 22.08 & 0.5797 & 0.1828 & 0.0378 \\
\rowcolor{gray!10}
$\text{SAIL-Caption-SA1B}_{\text{Subset}}$ & CN & 10,000 & 156.8 & 88.2 & 7.742 & 22.74 & 0.5872 & 0.1688 & 0.0359 \\

BLIP3-KALE~\citeyearpar{awadalla2024blip3} & CN & 235,125,090 & 66.16 & 73.2 & 21.08 & 42.83 & 0.8686 & 0.0930 & 0.7012 \\
\rowcolor{gray!10}
$\text{SAIL-Caption-KALE}_{\text{Subset}}$ & CN & 10,000 & 63.53 & 80.6 & 16.79 & 32.08 & 0.9107 & 0.0634 & 0.5768 \\

\bottomrule
\end{tabular}
}
\centering
\caption{
Statistics of SAIL-Caption and other opensource datasets. 
``Quality” refers to quality scores evaluated by human annotators.
We employ NLTK~\cite{bird2006nltk} and Jieba~\cite{sunchinese} to perform text segmentation and part-of-speech tagging for English and Chinese captions, respectively.
``Avg. Len.” stands for ``average length” and ``Uni.” denotes ``unique” items per sample.
Statistics of SAIL-Caption subsets are marked with \colorrect{mygray}. 
}
\label{tbl: sailcaption_statistics}
\vspace*{-2mm}
\end{table*}

\subsection{Pretrain}
During pretraining, we gradually open model parameters for larger-scale pretraining to develop SAIL-VL's visual understanding abilities.
We start from a randomly initialized multi-layer perceptron (MLP) module as the vision-to-language projector, and train it with approximately 131B tokens of detail caption and OCR data in the \textit{Pretrain-Alignment} stage. 
After warming up, we unlock the visual encoder of SAIL-VL for larger model capacity during the following \textit{Pretrain-Advance} stage, and train the model through approximately 524B tokens. 
Note that we do not use the entire SAIL-Caption dataset but a subset with an even distribution instead to ensure the diversity in data distribution.
For OCR data, we use several high-quality OCR datasets repeatedly instead of incorporating diverse but relatively low-quality data. 
The advantage of using repeated-yet-high-quality data is shown in Section~\ref{sec: 5_1}.
For SAIL-VL-8B, we allocate 20B- and 32B-token training budgets in the two pretraining stages for efficiency.




\subsection{SFT}
\label{sec: 2_2}
We train all parameters of SAIL-VL in a curriculum learning fashion with progressively higher-complexity training data in SFT stages. 
In the first \textit{SFT-Knowledge} stage, SAIL-VL learns basic instruction-following abilities and ingests world knowledge from Infinity-MM Stage2~\cite{gu2024infinity} data. 
During the subsequent \textit{SFT-Instruction} stage, we further optimize SAIL-VL towards enhanced visual instruction following capabilities with our customized 12M-sample high-quality visual instruction tuning dataset. 
For the final \textit{SFT-Preference} stage, we train SAIL-VL on a small amount of complex visual instruction tuning data, including LLaVA~\cite{li2024llava} SFT, Molmo Caption~\cite{deitke2024molmo}, and Infinity-MM Stage4~\cite{gu2024infinity} data, enabling SAIL-VL to tackle a wider range of complex instruction following tasks. 
We refer to Section~\ref{sec: 5_2} for detailed data distribution of the three stages.

\section{Towards Scalable VLM Training}

In this section, we introduce our scalable high-quality data construction pipeline and elaborate on the model performance scaling laws observed in both pretraining and SFT stages. 

\subsection{Scalable High-Quality Visual Understanding Data Construction}
\label{sec: 3_1}

Our scalable data construction pipeline is shown in Figure~\ref{fig: pipeline}, consisting of the following four steps. 

\paragraph{Data collection.}
\label{sec: 3_1_1}
We collect source data from a wide range of public image datasets to ensure data distribution diversity.
Our source datasets include LAION-COCO~\cite{schuhmann2022laion}, TextCaps~\cite{sidorov2020textcaps}, SA1B~\cite{kirillov2023segment}, and several other large-scale datasets. 

\paragraph{Reference data curation.}
\label{sec: 3_1_2}
We curate a small amount of reference data to train a compact VLM for efficient data annotation at scale. 
We first select a subset of source images with a balanced distribution, and then task GPT4-O-20240513~\cite{gpt4o} deployed by Azure to annotate detail captions.
Following previous works~\cite{yu2024capsfusion,hong2024cogvlm2}, alt-texts are provided if available for supplementary world knowledge and enhanced reference data quality. 

\paragraph{Captioner model training.}
Equipped with the high-quality reference data, we train an InternVL2-8B~\cite{internvl2} model on the reference data to generate high-quality data at scale, which is called SAIL-Captioner. 
Similarly, alt-texts are optionally included in the caption generation prompt, enabling SAIL-Captioner to perform both captioning and recaptioning tasks.

\paragraph{Scalable high-quality data construction.}
In the final stage, we deploy SAIL-Captioner with LMDeploy~\cite{2023lmdeploy} for large-scale detail caption data construction. 
We implement a multi-task, multi-node, and multi-processing asynchronous annotation pipeline, enabling flexible computation resource allocation. 



\begin{figure*}[t]
\centering
\resizebox{\textwidth}{37mm}{
\begin{tikzpicture}
\draw (0,0) node[inner sep=0] {\includegraphics[width=\columnwidth, trim={0cm 0cm 0cm 0cm}, clip]{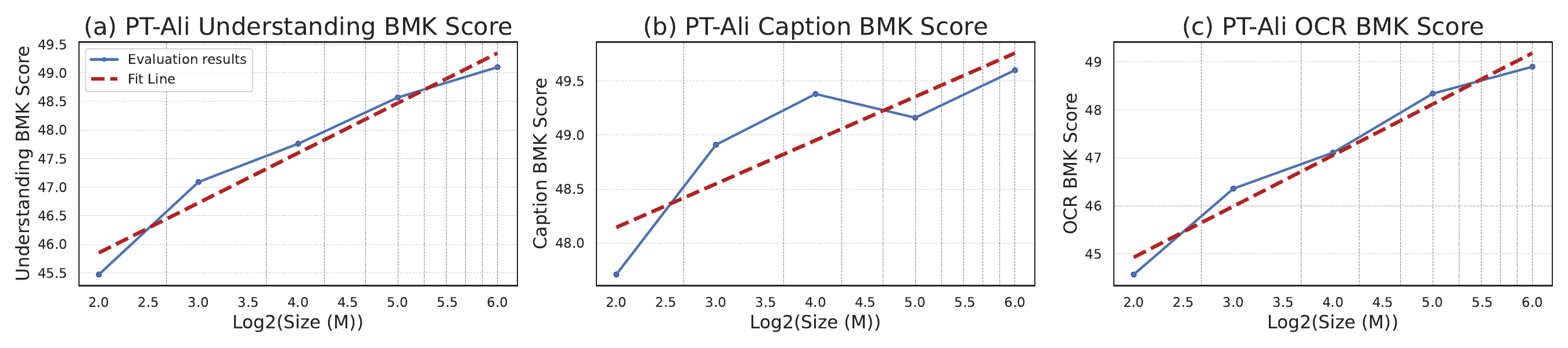}};
\end{tikzpicture}
}
\caption{
Scaling curves of SAIL-VL-2B's performance dynamics in the pretrain-alignment (PT-Ali) stage. 
We show model performance on all understanding benchmarks, caption tasks and OCR tasks, respectively.
``BMK Score” stands for average benchmark scores. 
}
\label{fig: ptali_scaling}
\vspace{-2mm}
\end{figure*}

\begin{figure*}[t]
\centering
\resizebox{\textwidth}{37mm}{
\begin{tikzpicture}
\draw (0,0) node[inner sep=0] {\includegraphics[width=\columnwidth, trim={0cm 0cm 0cm 0cm}, clip]{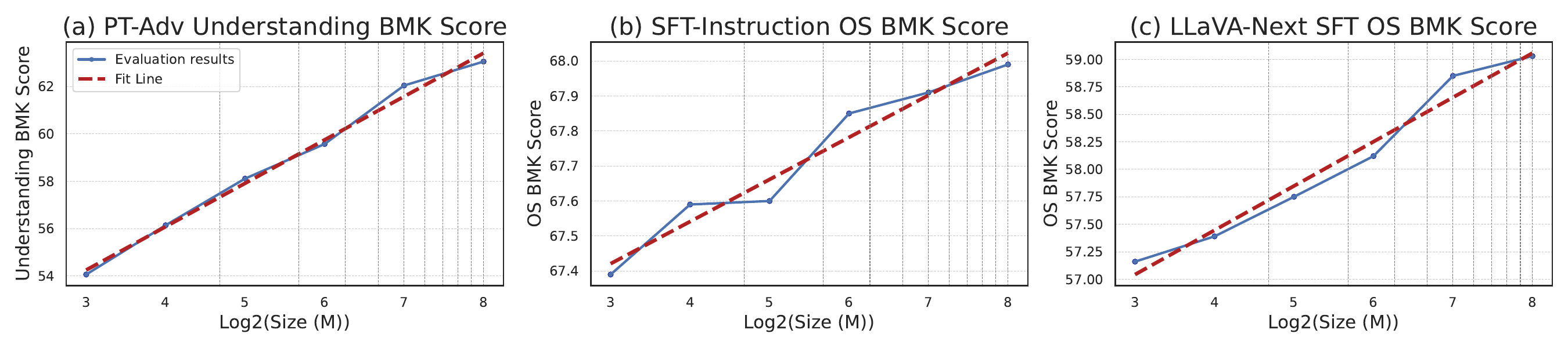}};
\end{tikzpicture}
}
\caption{
Scaling curves of SAIL-VL-2B's performance dynamics in the pretrain-advance (PT-Adv) stage. 
We show pretrained and SFT model performance on understanding benchmarks and OS (opensource) VLM benchmarks, respectively.
``BMK Score” stands for average benchmark scores. 
}
\label{fig: ptadv_scaling}
\vspace{-2mm}
\end{figure*}

\paragraph{SAIL-Caption.}
Equipped with the aforementioned data construction pipeline, we construct SAIL-Caption, a detail caption dataset with 300M image samples from various sources. 
To validate the data quality of SAIL-Caption, we randomly sample 10,000 cases from SAIL-Caption and other opensource caption datasets for comparison, and the statistics are shown in Table~\ref{tbl: sailcaption_statistics}. 
Results show that SAIL-Caption is not only of large quantity, but also demonstrates leading richness of visual elements, for example, unique n-grams, nouns, verbs, and adjectives in caption texts. 
These statistics indicate that SAIL-Caption encompasses more visual elements and exhibits greater linguistic diversity in caption texts. 
Moreover, SAIL-Caption receives higher quality scores from human annotators, surpassing existing opensource datasets by a large margin. 
We refer to Appendix~\ref{sec: appendix_sailcaption} for detailed caption quality evaluation procedure and SAIL-Caption showcases. 

\subsection{Scalable VLM Pretraining with High-Quality Visual Understanding Data}
In this part, we introduce the data size scaling laws observed in SAIL-VL-2B large-scale pretraining. 
For model checkpoints obtained at different pretraining steps, we conduct lightweight annealing training with 2M identically distributed data for improved convergence and evaluation stability. 

\subsubsection{Improving VLM Visual Understanding Performance via Data Size Scaling}
SAIL-VL-2B is trained through 131B and 524B tokens during the two pretraining stages, respectively, during which we investigate model performance dynamics.
To evaluate the visual understanding performance of SAIL-VL, we establish an evaluation suite which covers fundamental visual understanding tasks such as detail caption generation~\cite{dong2024benchmarking} and OCR detection~\cite{biten2022ocr,wang2020docstruct,Gupta16,kim2022donut}. 
Details can be found in Appendix~\ref{sec: appedix_understanding_bmk}.
As shown in Figure~\ref{fig: ptali_scaling}, SAIL-VL's visual understanding performance in each domain improves steadily in the pretrain-alignment stage. 
As the training data size scales up exponentially, the model performance exhibits a linear growth trend. 
We also show the understanding performance dynamics in the pretrain-advance stage. 
SAIL-VL's understanding benchmark scores improve markedly in this stage, which we attribute to the large capacity of the vision encoder optimized for visual understanding. 
In Figure~\ref{fig: ptadv_scaling} (a), a similar linear performance scaling curve is observed, unveiling a promising prospect to scale up VLM pretraining data sizes for improved model performance. 


\subsubsection{Generalizing Visual Understanding Abilities to Instruction Following Tasks}
To further investigate the effectiveness of SAIL-VL's large-scale pretraining, we conduct SFT with different data collections for pretrain-advance model checkpoints trained with different data sizes.

As shown in Figure~\ref{fig: ptadv_scaling} (b)(c), the overall performance dynamics of SFT models can be plotted as a near-linear curve on an exponential horizontal axis, exhibiting smooth data size scaling laws on opensource VLM benchmarks.
We conduct experiments with both our SFT-Instruction data and opensource LLaVA-Next SFT data.
Despite the different data composition and final benchmark scores, similar scaling curves can be observed in both experiment sets. 
We further investigate pretrained and SFT model performance correlation in Appendix~\ref{sec: appendix_ptsft_correlation}.

\begin{figure}[t]
\centering
\begin{tikzpicture}
\draw (0,0) node[inner sep=0] {\includegraphics[width=\columnwidth, trim={0cm 0cm 0cm 0cm}, clip]{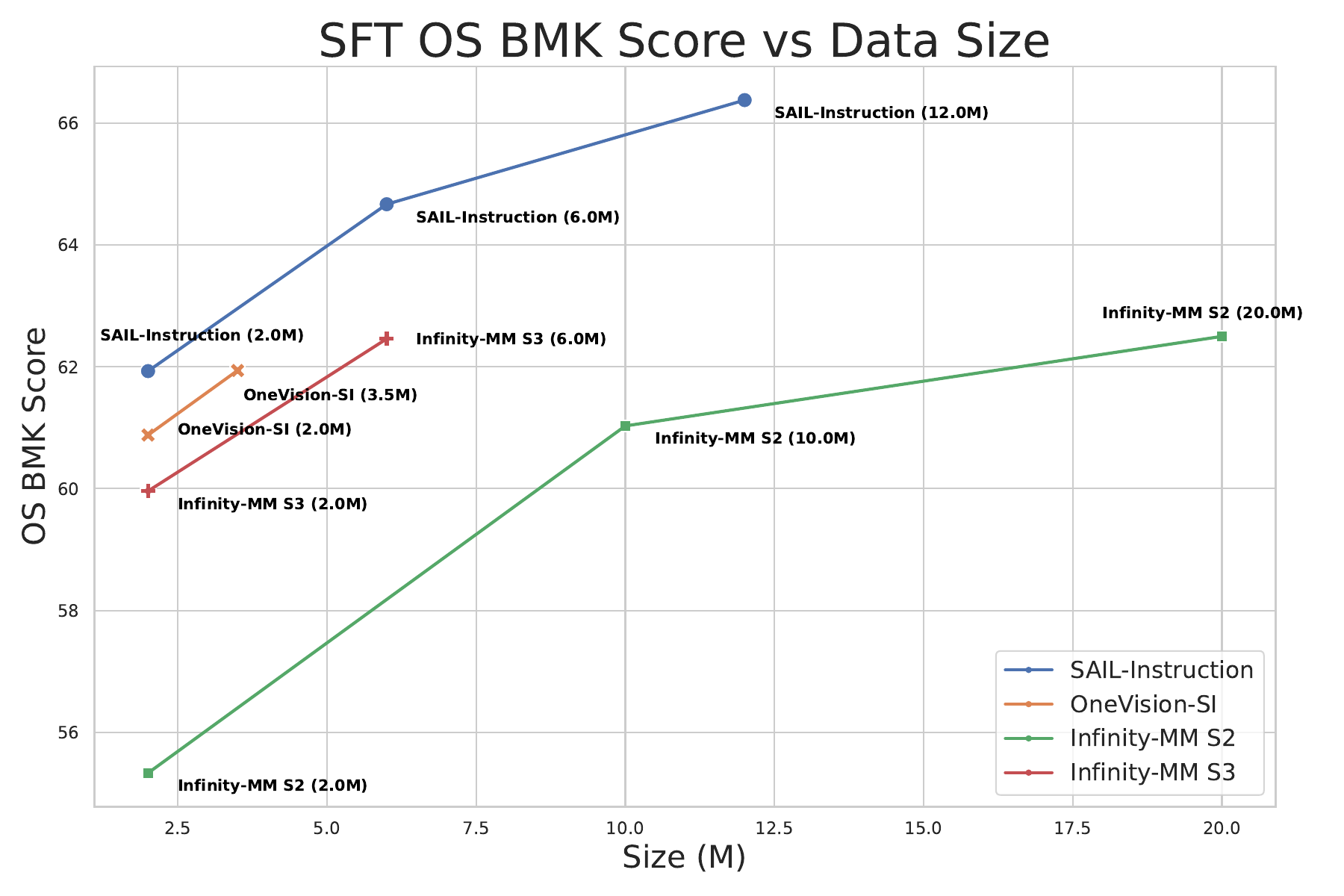}};
\end{tikzpicture}
\caption{
Scaling curves of model performance trained on our SAIL-Instruct dataset, LLaVA-OneVision~\cite{li2024llava} single image SFT data, and datasets from Infinity-MM~\cite{gu2024infinity}.
Model performance is shown as an average score across 18 benchmarks. 
}
\label{fig: sft_data_scale}
\vspace{-2mm}
\end{figure}

\subsection{Scaling up Visual Instruction Tuning}
Despite the abundance of publicly available visual instruction tuning data, high-quality training data is still scarce. 
We first introduce guidelines for our high-quality SFT data curation in Section~\ref{sec: 3_3_1}, and demonstrate model performance scaling laws of the curriculum SFT strategy in Section~\ref{sec: 3_3_2}. 

\subsubsection{High-Quality SFT Data Curation for Data Quantity Scaling}
\label{sec: 3_3_1}
In this part, we elaborate on the methodologies for visual instruction tuning data curation and demonstrate their effectiveness in SAIL-VL training. 

\begin{figure}[t]
\centering
\begin{tikzpicture}
\draw (0,0) node[inner sep=0] {\includegraphics[width=\columnwidth, trim={0cm 0cm 0cm 0cm}, clip]{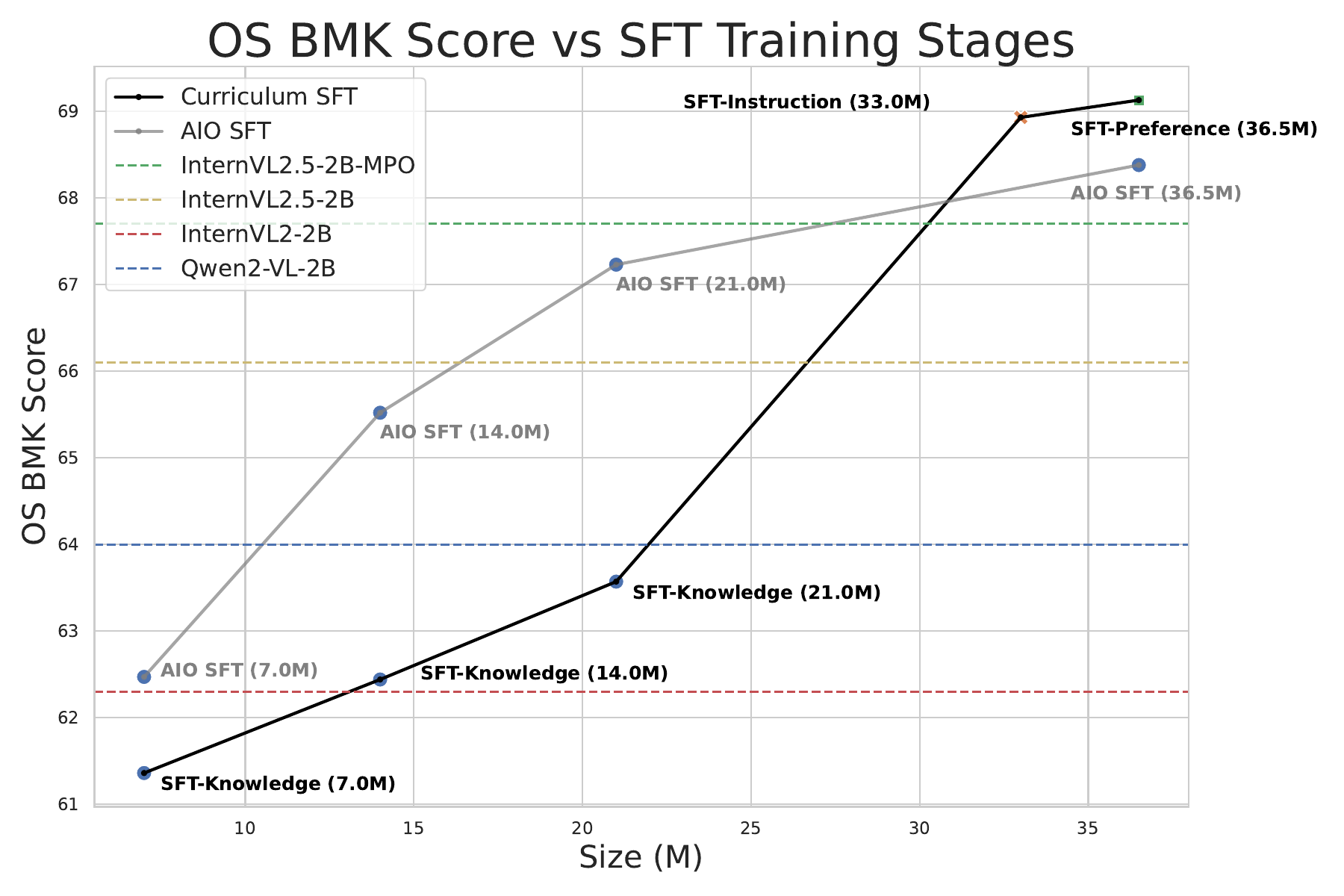}};
\end{tikzpicture}
\caption{
Model performance dynamics of the quality scaling and all-in-one (AIO) training strategy.
``AIO learning” incorporates all three-stage SFT data into a single training loop. 
Model performance is shown as an average score across 18 benchmarks. 
}
\label{fig: sft_quality_scaling}
\vspace{-2mm}
\end{figure}

\paragraph{High-quality visual instruction tuning dataset curation. }
To judge the quality of different SFT data collections efficiently, we start with the \textit{Quick Quality Evaluation} strategy.
This strategy assesses the quality of a given SFT data collection by training with its 2M-sample subset. 
The resulting model performance reflects the training data quality, enabling efficient data quality evaluation and comparison. 
In this strategy, we assume that models trained on different datasets maintain a consistent performance ranking across varying training data sizes.
This assumption is validated by experiment results shown in Figure~\ref{fig: sft_data_scale}.

\begin{table}[t]
\centering
\resizebox{0.45\textwidth}{!}{
\begin{tabular}{l | c c c c}
\toprule
SFT Stage & Size $\downarrow$ & Diff. $\uparrow$ & Comp. $\uparrow$ & Rel. $\uparrow$\\

\midrule

SFT-Knowledge & 21M & 1.90 & 2.44 & 3.94 \\
SFT-Instruction & 12M & 2.15 & 2.62 & 4.45 \\
SFT-Preference & 3.5M & 2.20 & 2.74 & 4.55 \\

\bottomrule
\end{tabular}
}
\caption{
Sizes and quality evaluation results of the three-stage SFT data. 
``Diff.”, ``Comp.”, and ``Rel.” stand for task difficulty, data complexity, and imaget-text relevance, respectively. 
}
\label{tbl: sft_data_statistics}
\end{table}

\begin{table*}[t]
\resizebox{\textwidth}{!}{
\begin{tabular}{ll | cc | cc | cc | cc | cc}
\toprule
\multicolumn{2}{l}{} &
\multicolumn{2}{|c}{Pretrain-Alignment} &
\multicolumn{2}{|c}{Pretrain-Advance} &
\multicolumn{2}{|c}{SFT-Knowledge} &
\multicolumn{2}{|c}{SFT-Instruction} &
\multicolumn{2}{|c}{SFT-Preference} \\
\cmidrule(lr{0.5em}){3-4}\cmidrule(lr{0.5em}){5-6}\cmidrule(lr{0.5em}){7-8}\cmidrule(lr{0.5em}){9-10}\cmidrule(lr{0.5em}){11-12}
&  & 2B & 8B & 2B & 8B & 2B & 8B & 2B & 8B & 2B & 8B \\
\midrule

\multirow{2}{*}{\rotatebox[origin=c]{90}{\footnotesize \textit{Vision}}} 
& Resolution & \multicolumn{2}{c|}{\footnotesize {$448\times\{\{1\times1\},...,\{2\times4\}\}$}} & \multicolumn{2}{c|}{\footnotesize {$448\times\{\{1\times1\},...,\{2\times4\}\}$}} & \multicolumn{2}{c|}{\footnotesize {$448\times\{\{1\times1\},...,\{2\times5\}\}$}} & \multicolumn{2}{c|}{\footnotesize {$448\times\{\{1\times1\},...,\{2\times5\}\}$}} & \multicolumn{2}{c}{\footnotesize {$448\times\{\{1\times1\},...,\{2\times5\}\}$}} \\
& \# Max Visual Token & \multicolumn{2}{c|}{2048} & \multicolumn{2}{c|}{2048} & \multicolumn{2}{c|}{2560} & \multicolumn{2}{c|}{2560} & \multicolumn{2}{c}{2560} \\

\midrule

\multirow{2}{*}{\rotatebox[origin=c]{90}{\footnotesize \textit{Data}}} 
& Data Composition & \multicolumn{2}{c|}{SAIL-Caption \& OCR} & \multicolumn{2}{c|}{SAIL-Caption \& OCR} & \multicolumn{2}{c|}{Curated VQA Data} & \multicolumn{2}{c|}{Curated VQA Data} & \multicolumn{2}{c}{Curated VQA Data} \\
& Dataset Size & 64M & 10M & 256M & 16M & \multicolumn{2}{c|}{21M} & \multicolumn{2}{c|}{12M} & \multicolumn{2}{c}{3.5M} \\

\midrule

\multirow{3}{*}{\rotatebox[origin=c]{90}{\footnotesize \textit{Training}}} 
& Trainable Module & \multicolumn{2}{c|}{Projector} & \multicolumn{2}{c|}{Vision \& Projector} & \multicolumn{2}{c|}{Full Model} & \multicolumn{2}{c|}{Full Model} & \multicolumn{2}{c}{Full Model} \\
& Trainable Parameter & 8.65M & 27.52M & 313M & 332M & 1.85B & 7.95B & 1.85B & 7.95B & 1.85B & 7.95B \\
& Batch Size & 1920 & 512 & 2048 & 2048 & \multicolumn{2}{c|}{512} & \multicolumn{2}{c|}{512} & \multicolumn{2}{c}{512} \\
& Learning Rate & $1\times10^{-4}$ & $1\times10^{-3}$ & $4\times10^{-5}$ & $4\times10^{-5}$ & $2\times10^{-5}$ & $1\times10^{-5}$ & $2\times10^{-5}$ & $1\times10^{-5}$ & $2\times10^{-5}$ & $1\times10^{-5}$ \\

\bottomrule
\end{tabular}
}
\centering
\caption{
Details of the training pipeline of SAIL-VL-2B and SAIL-VL-8B.
}
\label{tbl: hyperparam}
\vspace{-1mm}
\end{table*}

We then propose the \textit{Composition Evaluation} strategy to judge the quality of existing SFT data components. 
In composition evaluation, we start with existing SFT data collections, for example, LLaVA-OneVision~\cite{li2024llava}, Cauldron~\cite{laurenccon2024matters}. 
We then categorize the datasets based on their format and distribution, resulting in a series of data groups, including closed-form VQA, open-ended VQA, document VQA, math\&reasoning, and pure text QA data~\footnote{Closed-form and open-ended VQA data refer to natural image VQA data requiring specific choice answers and open-ended responses, respectively}. 
To optimize the proportion of different data components, we halve each data group and judge the quality of the resulting data collection with our quick quality evaluation method. 
Once the model performance improves, the downward adjustment of the data proportion is retained. 

For incoming datasets to be incorporated into the SFT data, we conduct \textit{Incremental Evaluation}. 
Each new dataset is included in the SFT data collection, with the resulting data quality evaluated via lightweight model training. 
Datasets improving the model performance are regarded as beneficial for the data quality, and are therefore incorporated into our data collection. 
We also incorporate datasets which maintain the model performance, as they help expand the data scale for improved results.

\paragraph{Data Quantity Scaling.}
We curate our SAIL-Instruction data collection (used in SFT-Instruction stage) with the methodologies described above. 
To validate its advantage in data quality, we train the SAIL-VL model with our SAIL-Instruction data and other opensource SFT data collections at varying data scales. 
As shown in Figure~\ref{fig: sft_data_scale}, the performance of SAIL-VL scales up stably as the model training proceeds, depicting a logarithmic performance scaling curve.  
Compared with other opensource SFT data collections, our SAIL-Instruction data achieves the highest model performance at every data point. 
It is also worth noticing that the performance ranking of models trained with different datasets remains consistent across the training process. 
This observation validates our quick quality evaluation method introduced above.

\subsubsection{Multi-stage Instruction Tuning for Data Complexity Scaling}
\label{sec: 3_3_2}
In this part, we introduce data complexity scaling, a curriculum learning strategy for VLM SFT for enhanced model performance.

\paragraph{Curriculum SFT with progressively improving data quality.}
As elaborated in Section~\ref{sec: 2_2}, we train SAIL-VL through three SFT stages, and the data collections used in later stages differ from previous ones in the following aspects: 
(1) Datasets are harder to collect and therefore of smaller quantity.
(2) Training tasks become increasingly challenging, and the questions in the training data are more difficult to answer.  
(3) Data complexity progressively increases, requiring more fine-grained understanding of the visual elements and in-depth reasoning.
We validate our design by quantifying the data distribution variance across the three stages via human evaluation. 
As shown in Table~\ref{tbl: sft_data_quality}, the task difficulty, data complexity, and image-text relevance increase monotonously across the three stages.
SFT data in the later stages is of higher overall quality, but is also more challenging for the model to learn from, which coincides with our curriculum SFT design. 
We refer to Appendix~\ref{sec: appendix_sft_data_eval} for the detailed definition of these data quality dimensions and the full instruction for human evaluation.





\paragraph{Data Complexity Scaling. }
To demonstrate the effectiveness of our curriculum SFT strategy with progressively higher-complexity data, we show model performance dynamics derived from the three SFT stages in comparison with an all-in-one (AIO) training strategy in Figure~\ref{fig: sft_quality_scaling}. 
The model trained with our curriculum SFT strategy exhibits a near-linear performance scaling curve across training stages, outperforming the logarithmic scaling curve of AIO training baseline.
This result validates the marked effectiveness of the curriculum SFT strategy. 
Training with small and high-complexity SFT data in later stages yields more promising performance scaling curves. 

\begin{table*}[t]
\resizebox{\textwidth}{57mm}{
\begin{tabular}{l | c c c c | c c c c}
\toprule
Benchmark & SAIL-VL & Qwen2-VL & InternVL2.5-MPO & DeepSeekVL-2 & SAIL-VL & Qwen2-VL & InternVL2.5-MPO & DeepSeekVL-2 \\

\cmidrule(lr{0.5em}){2-5}\cmidrule(lr{0.5em}){6-9}
 & \multicolumn{4}{c|}{2B Model} & \multicolumn{4}{c}{8B Model} \\

\midrule
\rowcolor{gray!15} \multicolumn{9}{c}{\textit{Overall Performance}} \\
\midrule
Opensource Average        & \textbf{69.1} & 64.4 & \underline{67.7} & 67.0 & \textbf{74.5} & 73.0 & \underline{74.3} & 72.7  \\
\quad General VQA        & \underline{60.4} & 58.3 & \textbf{63.1} & 59.4 & 68.3 & \underline{68.5} & \textbf{71.2} & 66.8  \\
\quad OCR VQA        & \textbf{75.9} & 72.5 & 71.1 & \underline{74.4} & \textbf{79.8} & \underline{79.6} & 76.3 & 79.0  \\
\quad Math\&Knowledge        & \textbf{79.0} & 59.0 & \underline{75.3} & 71.3 & \textbf{83.3} & 71.0 & \underline{83.2} & 79.0  \\
\quad Hallucination        & \textbf{66.2} & 62.9 & \underline{64.5} & 63.6 & \underline{68.7} & 67.5 & \textbf{69.7} & 65.3 \\

\midrule
\rowcolor{gray!15} \multicolumn{9}{c}{\textit{General VQA}} \\
\midrule
MMStar~\citeyearpar{chen2024we}        & \textbf{55.1} & 46.3 & \underline{54.3} & 49.9 & \underline{64.2} & 58.3 & \textbf{65.3} & 57.7  \\
$\text{MMBench}_{\text{DEV}}$~\citeyearpar{liu2024mmbench}        & \underline{72.4} & 68.8 & \textbf{72.5} & 68.3 & 79.5 & 79.5 & \textbf{83.3} & 78.1  \\
$\text{MMMU}_{\text{VAL}}$~\citeyearpar{yue2024mmmu}        & \underline{40.1} & 39.9 & \textbf{41.2} & 39.6 & 48.2 & \underline{50.9} & \textbf{52.8} & 47.6  \\
MME~\citeyearpar{fu2023mme}        & \underline{1969} & 1923 & \textbf{2123} & 1910 & 2244 & \underline{2321} & \underline{2321} & 2149  \\
$\text{SEEDBench}_{\text{IMG}}$~\citeyearpar{li2023seed}        & \textbf{74.7} & 72.0 & \underline{73.2} & 72.5 & 75.5 & 75.3 & \textbf{76.9} & \underline{76.8}  \\
RealWorldQA~\citeyearpar{grok15_v}        & \underline{63.8} & 60.9 & 60.7 & \textbf{64.8} & \textbf{71.9} & 69.7 & 70.2 & 70.2  \\
MMVet~\citeyearpar{yu2024mm}        & 46.1 & 51.2 & \textbf{64.0} & \underline{52.8} & 58.3 & \underline{62.6} & \textbf{66.8} & 60.3  \\

\midrule
\rowcolor{gray!15} \multicolumn{9}{c}{\textit{OCR VQA}} \\
\midrule
$\text{AI2D}_{\text{TEST}}$~\citeyearpar{kembhavi2016diagram}        & \textbf{79.0} & 72.3 & \underline{75.3} & 74.6 & \underline{83.7} & 82.9 & \textbf{84.1} & 82.0  \\
$\text{DocVQA}_{\text{VAL}}$~\citeyearpar{mathew2021docvqa}        & \textbf{89.2} & \underline{88.7} & 87.8 & 88.6 & 92.2 & \textbf{93.7} & 92.1 & \underline{92.3}  \\
$\text{InfoVQA}_{\text{VAL}}$~\citeyearpar{mathew2022infographicvqa}        & \textbf{67.2} & 63.4 & 61.6 & \underline{63.8} & 75.2 & \underline{75.9} & \textbf{76.2} & 72.5  \\
$\text{ChartQA}_{\text{TEST}}$~\citeyearpar{masry2022chartqa}        & \underline{81.0} & 70.6 & 70.9 & \textbf{81.2} & \underline{84.6} & 81.6 & 77.6 & \underline{84.6}  \\
$\text{TextVQA}_{\text{VAL}}$~\citeyearpar{singh2019towards}        & 75.7 & \underline{78.8} & 77.2 & \textbf{80.5} & 77.7 & \textbf{83.8} & 79.2 & \underline{83.3}  \\
$\text{OCRVQA}_{\text{TEST}}$~\citeyearpar{mishraICDAR19}        & \textbf{58.5} & \underline{54.3} & 40.0 & 51.4 & \textbf{61.4} & \underline{56.2} & 36.7 & 54.5  \\
OCRBench~\citeyearpar{liu2024ocrbench}        & 806 & 794 & \textbf{846} & \underline{808} & \underline{835} & 833 & \textbf{880} & 834  \\

\midrule
\rowcolor{gray!15} \multicolumn{9}{c}{\textit{Math\&Knowledge}} \\
\midrule
$\text{MathVista}_{\text{MINI}}$~\citeyearpar{lu2023mathvista}        & \textbf{62.8} & 45.0 & \underline{55.3} & 54.5 & \underline{68.4} & 57.3 & \textbf{68.5} & 61.8  \\
$\text{ScienceQA}_{\text{VAL}}$~\citeyearpar{saikh2022scienceqa}        & \underline{95.3} & 73.0 & \underline{95.3} & 88.1 & \textbf{98.2} & 84.6 & \underline{97.9} & 96.2  \\


\midrule
\rowcolor{gray!15} \multicolumn{9}{c}{\textit{Hallucination}} \\
\midrule
HallusionBench~\citeyearpar{guan2024hallusionbench}        & \textbf{45.7} & 38.3 & \underline{39.2} & 38.4 & \textbf{52.2} & 48.5 & \underline{50.3} & 41.2  \\
POPE~\citeyearpar{Li-hallucination-2023}        & 86.7 & 87.6 & \textbf{89.8} & \underline{88.8} & 85.2 & 86.5 & \underline{89.1} & \textbf{89.4}  \\

\bottomrule
\end{tabular}
}
\centering
\vspace{0mm}
\caption{
Evaluation results of SAIL-VL and other opensource VLM with comparable sizes.
``Opensource average” includes all opensource benchmarks listed in the table. 
Bold numbers indicate the best performance among models of comparable sizes, while underlined ones are those ranked as the second. 
}
\label{tbl: main}
\end{table*}

\section{Experiments}

\subsection{Experiment Setup}
\label{sec: 4_1}

\paragraph{Model Training.}
We start from InternViT-300M~\cite{chen2023internvl}, Qwen2.5-2B and Qwen2.5-7B~\cite{qwen2.5} for model training. 
Detailed model training recipes are elaborated in Table~\ref{tbl: hyperparam}.
As training progresses, the input image resolution gradually increases, with a $2\times2$ pixel shuffle~\cite{chen2024far} module employed in the projector, maintaining a balance between efficiency and performance.
For SAIL-VL-8B, we use smaller batch sizes and larger learning rates in pretraining stages to improve training efficiency. 
During SFT stages, the 8B model is trained with a smaller learning rate, mitigating the instability in full model training with larger LLMs.

\paragraph{Baselines.}
We compare our SAIL-VL models with previous SOTA VLM baselines of comparable sizes, including Qwen2-VL~\cite{wang2024qwen2}, InternVL2.5-MPO~\cite{chen2024expanding}, DeepSeekVL-2\cite{wu2024deepseekvl2mixtureofexpertsvisionlanguagemodels}, etc. 
Evaluation results against more existing baseline models are shown in Appendix~\ref{sec: appendix_exp_detail_result}.

\paragraph{Evaluation.}
We evaluate SAIL-VL and baseline VLMs on a series of widely used benchmarks, including General VQA, OCR VQA, Math\&Knowledge, and Hallucination.
These categories cover VQA tasks on natural images/videos, OCR-related documents, as well as those involving complicated reasoning abilities and world knowledge to tackle. 
We use a customized version of VLMEvalKit~\cite{duan2024vlmevalkit} for evaluations.


\subsection{Benchmark Results}

\paragraph{SAIL-VL-2B ourperforms previous SOTA VLMs with comparable sizes significantly.}
We list the performance of SAIL-VL along with other opensource VLMs in Table~\ref{tbl: main}. 
As the results show, SAIL-VL-2B outperforms previous SOTA VLMs by a large margin, scoring $1.4$ ($2.06\%\uparrow$) higher average performance than InternVL2.5-MPO-2B.
SAIL-VL-2B achieves new SOTA performance on 3 out of 4 subfields except for General VQA.
We attribute it to the instability lying in benchmarks requiring long text generation, such as MMVet. 

\paragraph{SAIl-VL-8B achieves leading performance over opensource baselines.}
As shown in Table~\ref{tbl: main}, SAIL-VL-8B also achieves leading visual comprehension performance over Qwen2-VL, DeepSeekVL-2, and even InternVL2.5-MPO-8B, which requires an additional reinforcement learning stage in model training. 
We admit the shrunk performance advantage of SAIL-VL-8B over SOTA baselines, which may be caused by the relatively small data sizes used for model training. 
We take these results as an early attempt for larger VLM training, and more competitive large VLMs will be released in our SAIL-VL series. 

\section{Analysis}

\subsection{Pretrain Data Quality Determines Pretrained Model Performance}
\label{sec: 5_1}
We explore pretraining SAIL-VL-2B with varying data quality.
Specifically, we conduct lightweight 16B-token training in the pretrain-advance stage, starting from the model checkpoint after the same alignment pretraining. 
We fix the data distribution across different data types, and modify data composition with varying-quality data. 

As shown in Table~\ref{tbl: pretrain_data_ablation}, the model trained with SAIL-Caption achieves significantly higher performance than those trained on other opensource caption datasets, which is consistent with data quality evaluation results as shown in Appendix~\ref{tbl: sailcaption_qualityeval}. 
It is also worth noticing that the model trained with repeated-yet-high-quality OCR data yields better results than incorporating diverse but relatively low-quality data for model training. 
We attribute this result to our frozen-LLM pretraining setting, which mitigates the potential overfitting problem lying in repeated training data.

\begin{table}[t]
\resizebox{0.45\textwidth}{13mm}{


\begin{tabular}{l l c c c}
\toprule
Caption Data & OCR Data & Overall & Caption & OCR \\
\midrule

SAIL-Caption & HQ & 54.36 & 51.80 & 55.38  \\
SA1B-QwenVL-Caption & HQ & 48.43 & 39.57 & 51.97  \\
DataComp-LLaVA-Caption & HQ & 49.08 & 42.70 & 51.63  \\
BLIP3-KALE & HQ & 53.06 & 46.00 & 55.89  \\

SAIL-Caption & HQ+LQ & 52.13 & 51.22 & 52.50  \\
SAIL-Caption & HQ (RP) & 54.05 & 52.63 & 54.62  \\

\bottomrule
\end{tabular}
}
\centering
\caption{
Visual understanding performance of model checkpoints pretrained with different data sources. 
We report models performance on our visual understanding benchmarks.
``HQ”, ``LQ”, and ``RP” indicates high-quality, low-quality, and repeated data, respectively. 
}
\label{tbl: pretrain_data_ablation}
\end{table}






\subsection{SFT Data Quality Analysis}
\label{sec: 5_2}
To further validate our data quality evaluation results shown in Table~\ref{tbl: sailcaption_statistics}, we select 2M-sample subsets from each SFT stage to train the pretrained SAIL-VL-2B model.
Performance evaluation results are shown in Table~\ref{tbl: sft_data_quality}. 
A significant performance advantage is observed in the model trained with SFT-Instruction data collection, validating the effectiveness of the proposed data curation methods. 
This result coincides with the data quality evaluation results given in Table~\ref{tbl: sailcaption_statistics}, where SFT-Instruction data collection exhibits advanced task difficulty, data complexity, and image-text relevance. 
It is also worth noticing that despite the improved data quality of the SFT-Preference data, it fails to further improve model performance in Table~\ref{tbl: sft_data_quality}. 
We attribute it to its excessively high data complexity, which may hinder effective model learning. 
This observation further validates the proposed curriculum VLM SFT strategy as discussed in Section~\ref{sec: 3_3_2}.

\begin{table}[t]
\centering
\resizebox{0.48\textwidth}{!}{
\begin{tabular}{l|ccccc}
\toprule
Training Data & Overall & General & OCR & Math. & Hall. \\
\midrule

SFT-Knowledge & 57.8 & 53.2 & 60.9 & 56.9 & 63.9 \\
SFT-Instruction & 61.9 & 55.8 & 67.1 & 65.4 & 61.7 \\
SFT-Preference & 61.3 & 57.1 & 65.8 & 59.5 & 61.3 \\

\bottomrule
\end{tabular}
}
\caption{
Performance evaluation results of models trained with SFT data from each stage. 
We denote ``Math.” as Math~\&Knowledge benchmarks in evaluation. 
``Hall.” denotes Hallucination benchmarks as defined in Section~\ref{sec: 4_1}.
}
\label{tbl: sft_data_quality}
\end{table}

\section{Related Works}

\subsection{Visual Understanding Data}
Visual understanding data consists of vision modality contents and corresponding language depictions, and is regarded as the keystone to various vision and language model applications. 
Whether it is representation learning models like CLIP and its derivatives~\cite{radford2021learning,jia2021scaling,shen2022k,cherti2023reproducible,fang2023data}, generative models~\cite{wang2022image,li2021align,bao2022vlmo,yu2022coca,li2023blip}, or recent vision language models~\cite{li2023blip,liu2024visual,internvl2,bai2023qwen}, all of these methods are built upon large scale high-quality visual understanding data. 
LAION~\cite{schuhmann2021laion,schuhmann2022laion}, TaiSu~\cite{liu2022taisu}, Coyo~\cite{kakaobrain2022coyo-700m}, DataComp~\cite{gadre2024datacomp}, and etc. provide relatively low-quality alt-texts paired with source images.
Subsequent works such as ShareGPT4V~\cite{chen2023sharegpt4v} and ALLaVA~\cite{chen2024allava} annotate small scale high-quality caption data with powerful VLM APIs. 
To produce high-quality detail caption data at scale, CapsFusion~\cite{yu2024capsfusion}, World2Seq~\cite{wang2024world}, CAPTURE~\cite{dong2024benchmarking}, SA1B-Recaption~\cite{sa1bqwen}, DataComp-Recaption~\cite{li2024if}, and BLIP3-KALE~\cite{awadalla2024blip3} employ recaptioner models for efficient data annotation.
The resulting datasets are widely used in recent VLM research. 



\subsection{Vision Language Model Pretrain}
VLM pretraining benefits from higher-quality and larger-scale visual understanding data effectively.
Previous works, such as BLIP2~\cite{li2023blip} and LLaVA~\cite{liu2024visual}, pretrain the model with relatively low-quality caption datasets~\cite{li2023blip}.
Subsequent works, such as MiniCPM-V~\cite{yao2024minicpm}, InternVL~\cite{chen2024far,internvl2,chen2024expanding}, and QwenVL~\cite{bai2023qwen,wang2024qwen2} series, explore expanding high-quality visual understanding data sizes to improve model performance. 
In this work, we further reveal model performance dynamics w.r.t. SFT data quality and size, which are largely unexplored in previous works.

\subsection{Visual Instruction Tuning}
LLaVA~\cite{liu2024visual} first defines visual instruction tuning and provides a baseline for VLM SFT data curation. 
Subsequent LLaVA series models~\cite{liu2023improved,liu2024llavanext,li2024llava} refine the visual instruction tuning datasets and achieve significantly better model performance.
BLIP3~\cite{xue2024xgen} incorporates image-text interleaved data into visual instruction tuning, while CogVLM~\cite{hong2024cogvlm2}, InternVL~\cite{chen2024far,internvl2,chen2024expanding} and QwenVL series~\cite{wang2024qwen2} models explore using video question answering data for VLM SFT.
In this paper, we elaborate the guidelines for the design of visual instruction datasets, providing valuable references for VLM training.

\section{Conclusions}
In this work, we introduce SAIL-VL, an opensource vision language model series with SOTA performance.
We propose a scalable caption data construction pipeline and curate SAIL-Caption, a large-scale caption dataset with the highest quality among opensource alternatives.
Equipped with SAIL-Caption, we conduct large-scale pretraining with up to 655B tokens and demonstrate that even compact VLMs can benefit from scaled up training data size.  
We further present data size scaling laws that SAIL-VL's visual comprehension performance improves logarithmically as training data size increases. 
For visual instruction tuning stages, we elaborate on several key guidelines for high-quality SFT data curation, guided by which we curate our SFT-Instrcution dataset, a high-quality SFT data collection exhibiting improved model performance scaling curves than opensource alternatives during model training. 
The phased SFT strategy used in SAIL-VL SFT further improves the scaling curves from logarithmic to near-linear. 
We evaluate SAIL-VL on 18 opensource VLM benchmarks, and our model outperforms existing VLMs of comparable sizes consistently either in overall performance or domain-specific abilities, depicting promising prospects in real-world applications.


\section{Limitations}
Despite the leading performance of SAIL-VL among VLMs of comparable sizes, we acknowledge the potential insights that could be gained from experimenting with larger models.
We intend to explore this avenue in future work to enhance the robustness of the presented data size scaling laws and other findings.
Additionally, our exploration of data size scaling laws has been confined to a specific data magnitude. 
Although model performance is observed to be saturating at this data quantity, it remains uncertain whether there is room for further improvement under optimized training settings.

We also point out that although SAIL-VL's training process is designed carefully, models may generate hallucinated, biased, or harmful information under certain circumstances, which will be further discussed and mitigated in our future works.



\bibliography{custom}

\clearpage
\appendix

\section{Authorship and Credit Attribution}
\paragraph{Data Construction}
\textit{Hongyuan Dong}, \textit{Weijie Yin}

\paragraph{Pretraining}
\textit{Hongyuan Dong}, \textit{Weijie Yin}

\paragraph{SFT}
\textit{Zijiang Kang}, \textit{Weijie Yin}

\paragraph{Evaluation}
\textit{Weijie Yin}

\paragraph{Project Lead}
\textit{Xiao Liang}, \textit{Chao Feng}, and \textit{Jiao Ran}









\section{SAIL-VL Model Card}
We provide a simplified model card for the proposed SAIL-VL-2B and SAIL-VL-8B model. 

\begin{table}[h]
\resizebox{0.46\textwidth}{!}{
\begin{tabular}{l | cc}
\toprule
& SAIL-VL-2B & SAIL-VL-8B \\
\midrule
\# Parameter & 1.85B & 7.95B \\
LLM & Qwen2.5-1.5B & Qwen2.5-7B \\
Vision Encoder & \multicolumn{2}{c}{InternViT-300M} \\
Resolution & \multicolumn{2}{c}{\footnotesize {$448\times\{\{1\times1\},...,\{2\times5\}\}$}} \\
Adapter & \multicolumn{2}{c}{2-layer MLP} \\
Token Merge & \multicolumn{2}{c}{2$\times$2} \\
\# Max Visual Token & \multicolumn{2}{c}{2560} \\
\bottomrule
\end{tabular}
}
\centering
\caption{
SAIL-VL-2B and SAIL-VL-8B model card.
}
\label{tbl: model_card}
\vspace{-1mm}
\end{table}

\section{SAIL-VL Showcases}
\label{sec: appendix_sailvl_showcase}

\begin{figure*}[h!]
\centering
\resizebox{0.95\textwidth}{!}{
\begin{tikzpicture}
\draw (0,0) node[inner sep=0] {\includegraphics[width=\columnwidth, trim={0.1cm 13cm 0cm 0cm}, clip]{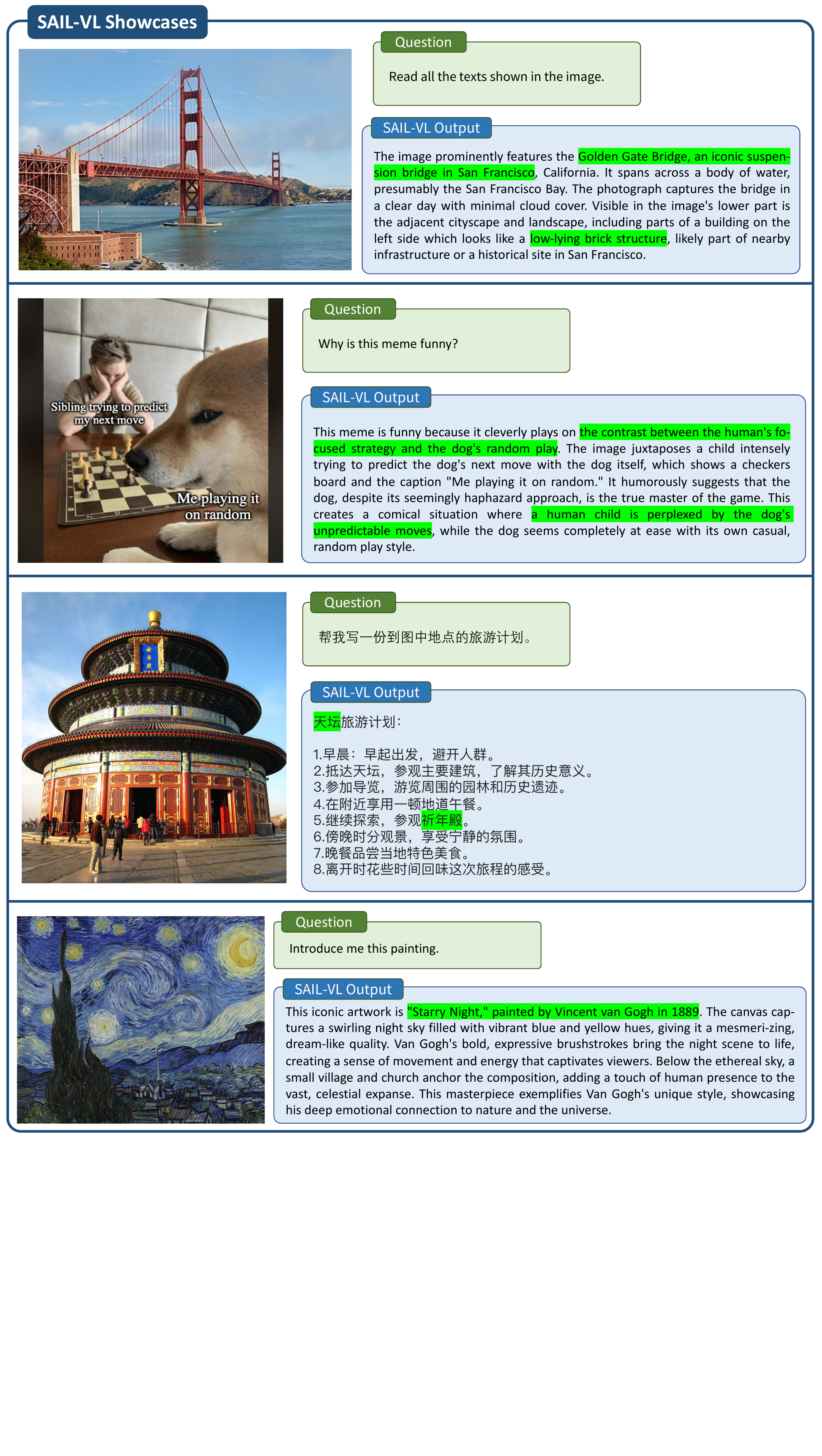}};
\end{tikzpicture}
}
\caption{
    SAIL-VL-8B showcases. 
    We include both English and Chinese queries with various input images. 
}
\label{fig: sailvl_showcase}
\end{figure*}

We task SAIL-VL-8B to tackle vision-based language interactions in Figure~\ref{fig: sailvl_showcase}. 
The images are selected from the internet, and input questions are set to cover common queries in real-world human interactions.
SAIL-VL demonstrates marked capabilities in language interactions in both English and Chinese. 
It also exhibits marked visual comprehension abilities for various input visual element types.
Our model recognizes famous landmarks, buildings, and artworks, demonstrating a vast reservoir of world knowledge.
It is also worth noticing that SAIL-VL also performs well in meme understanding. 
It not only perceives the visual elements accurately, but also points out the contrast that makes the meme humorous, exhibiting powerful visual comprehension and language interaction abilities. 


\section{SAIL-Caption}
\label{sec: appendix_sailcaption}

\subsection{Caption Data Quality Assessment}
\label{sec: appendix_sailcaption_quality}

\begin{table}[t]
\small
\resizebox{0.5\textwidth}{10mm}{
\begin{tabular}{l c c c c}
\toprule
Dataset & Language & Captioner & GPT Eval & Human Eval
\\
\midrule

$\text{DataComp-LLaVA-Caption}$ & EN & LLaVA-Captioner & 51.14 & 70.0 \\
\rowcolor{gray!10}
$\text{SAIL-Caption-DataComp}$ & EN & SAIL-Captioner & \textbf{61.50} & \textbf{87.2} \\

$\text{SA1B-QwenVL-Caption}$ & CN & QwenVL-Captioner & 63.82 & 74.6 \\
\rowcolor{gray!10}
$\text{SAIL-Caption-SA1B}$ & CN & SAIL-Captioner & \textbf{71.36} & \textbf{88.2} \\

$\text{BLIP3-KALE}$ & EN & 2B VLM & 59.88 & 73.2 \\
\rowcolor{gray!10}
$\text{SAIL-Caption-KALE}$ & EN & SAIL-Captioner & \textbf{61.50} & \textbf{80.6} \\

\bottomrule
\end{tabular}
}
\centering
\caption{
Data quality evaluation results of SAIL-Caption and other opensource caption datasets. 
The evaluation results of our SAIL-Captioner are marked with \colorrect{mygray}. 
The quality scores from GPT and human evaluation are rescaled to [0, 100] for simplicity. 
}
\label{tbl: sailcaption_qualityeval}
\end{table}
\textbf{}

\paragraph{Datasets. }
To evaluate the caption data quality of SAIL-Caption and opensource alternatives, we curate an evaluation subset for each recaption dataset as the test set. 
Specifically, we randomly select 500 samples from three recaption datasets listed in Table~\ref{tbl: sailcaption_qualityeval}.
SA1B-QwenVL-Caption employs a finetuned QwenVL~\cite{qwenvl_captioner} model to annotate Chinese dense captions on the SA1B dataset.
DataComp-LLaVA-Caption, on the other hand, trains a customized version of LLaVA~\cite{liu2023improved} model to perform annotation. 
BLIP3-KALE tasks CogVLM-18B~\cite{Wang2023CogVLMVE} and Mistral-8B-Instruct~\cite{jiang2024mixtral} to generate knowledge-augmented detail captions, and then distill this pipeline into a 2B VLM to annotate the DataComp-1B~\cite{gadre2024datacomp} dataset. 

We instruct Azure GPT4O-20240513~\cite{gpt4o} to generate ground truth captions for the curated evaluation sets.
We also use our SAIL-Captioner to generate detail captions on the evaluation sets. 
Captions from the original recaption dataset and those generated by our SAIL-Captioner model are then compared with the ground truth ones for evaluation.

\begin{figure*}[h!]
\centering
\resizebox{\textwidth}{!}{
\begin{tikzpicture}
\draw (0,0) node[inner sep=0] {\includegraphics[width=\columnwidth, trim={0.25cm 22cm 0.25cm 0cm}, clip]{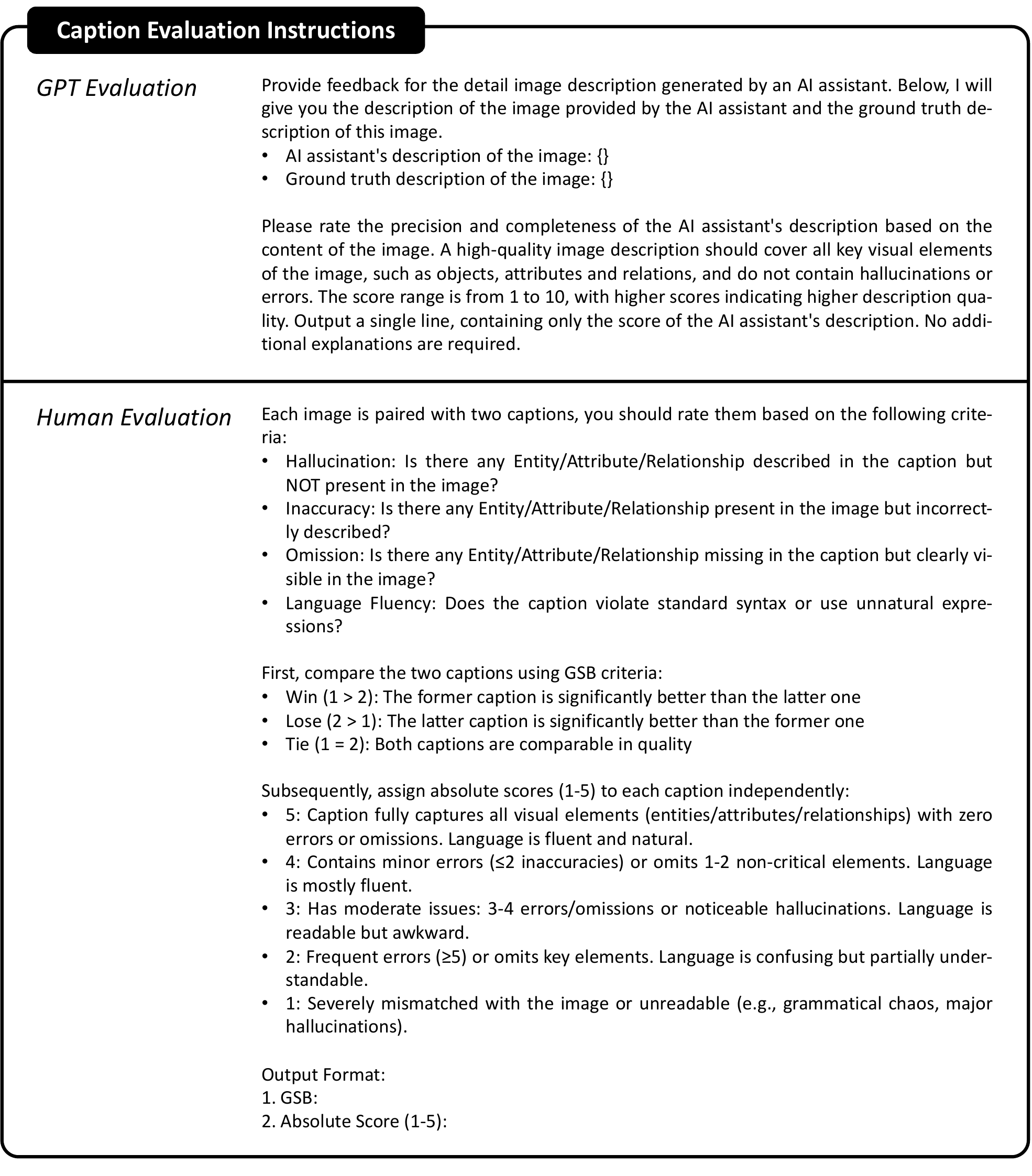}};
\end{tikzpicture}
}
\caption{
GPT and human evaluation prompts for SAIL-Caption and other opensource caption datasets. 
}
\label{fig: captioneval_prompt}
\vspace{-3mm}
\end{figure*}

\paragraph{GPT Evaluation. }
We first conduct GPT evaluation for efficiency. 
We feed the candidate captions and ground truth ones to Azure GPT4O API, and ask the model to judge the candidate caption quality based on the precision and recall of visual elements. 
The detailed prompt used for GPT evaluation is shown in Figure~\ref{fig: captioneval_prompt}. 

\paragraph{Human Evaluation. }
To further validate the advantage of the data quality of our SAIL-Caption dataset, we also task human experts to evaluate the data quality. 
We randomly select a 100-sample subset from each dataset and instruct the annotators to judge candidate caption quality based on the original image. 
As shown in Figure~\ref{fig: captioneval_prompt}, human annotators are given two candidate captions from both the baseline dataset and SAIL-Caption simultaneously. 
The experts are required to provide quality scores for the given captions, as well as a Good-Same-Bad (GSB) judgment reflecting more fine-grained data quality differences. 

In the inspection of 10\% annotated samples, we observe a 95\%+ accuracy, verifying the reliability of our human evaluation results.

\paragraph{Evaluation Results. }
As shown in Table~\ref{tbl: sailcaption_qualityeval}, SAIL-Caption-DataComp, SAIL-Caption-SA1B, and SAIL-Caption-KALE achieve significantly higher scores than previous baseline datasets in both GPT and human evaluation. 
These results demonstrate the leading performance of our SAIL-Captioner model and the advantage in SAIL-Caption's data quality. 

\begin{figure}[th]
\centering
\resizebox{0.46\textwidth}{!}{
\begin{tikzpicture}
\draw (0,0) node[inner sep=0] {\includegraphics[width=\columnwidth, trim={0cm 0cm 0cm 0cm}, clip]{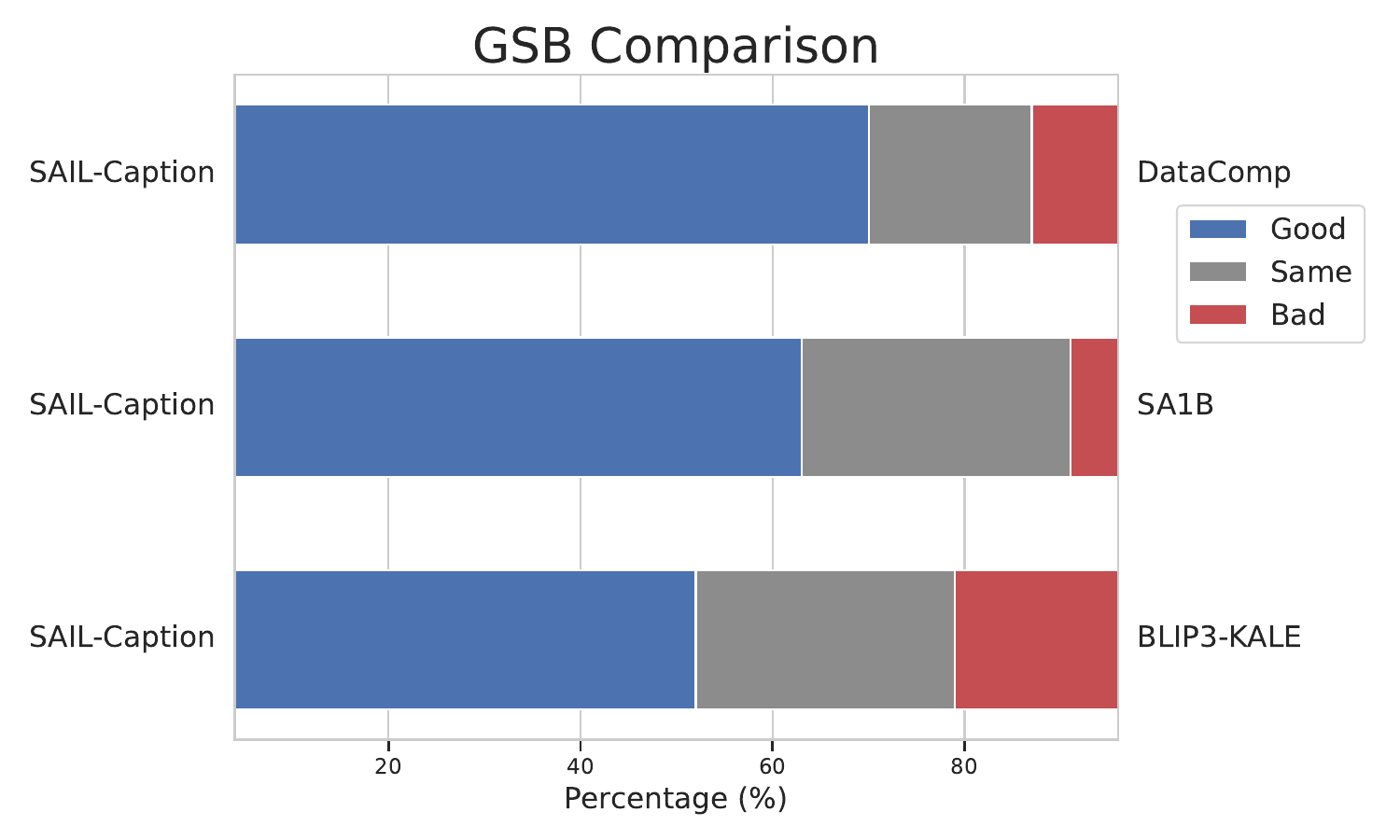}};
\end{tikzpicture}
}
\caption{
GPT and human evaluation prompts for SAIL-Caption and other opensource caption datasets. 
}
\label{fig: captioneval_gsb}
\end{figure}

We also show the GSB evaluation results in Figure~\ref{fig: captioneval_gsb}. 
The GSB comparison reflects more fine-grained caption quality differences in candidate captions than a single rating. 
In the GSB evaluation, SAIL-Captioner achieves 87\%, 91\%, and 79\% win+tie rates against SA1B-QwenVL-Caption, DataComp-LLaVA-Caption, and BLIP3-KALE, respectively, exhibiting marked quality advantages. 

We attribute the leading performance of SAIL-Captioner to the simple-yet-effective data distillation pipeline. 
SAIL-Captioner develops visual understanding abilities effectively from reference data annotated by powerful VLM APIs, enabling large-scale high-quality data generation with limited resources.

\subsection{SAIL-Caption Showcases}
\label{sec: appendix_sailcaption_showcase}

\begin{figure*}[h!]
\centering
\resizebox{\textwidth}{230mm}{
\begin{tikzpicture}
\draw (0,0) node[inner sep=0] {\includegraphics[width=\columnwidth, trim={0.25cm 15cm 1cm 0cm}, clip]{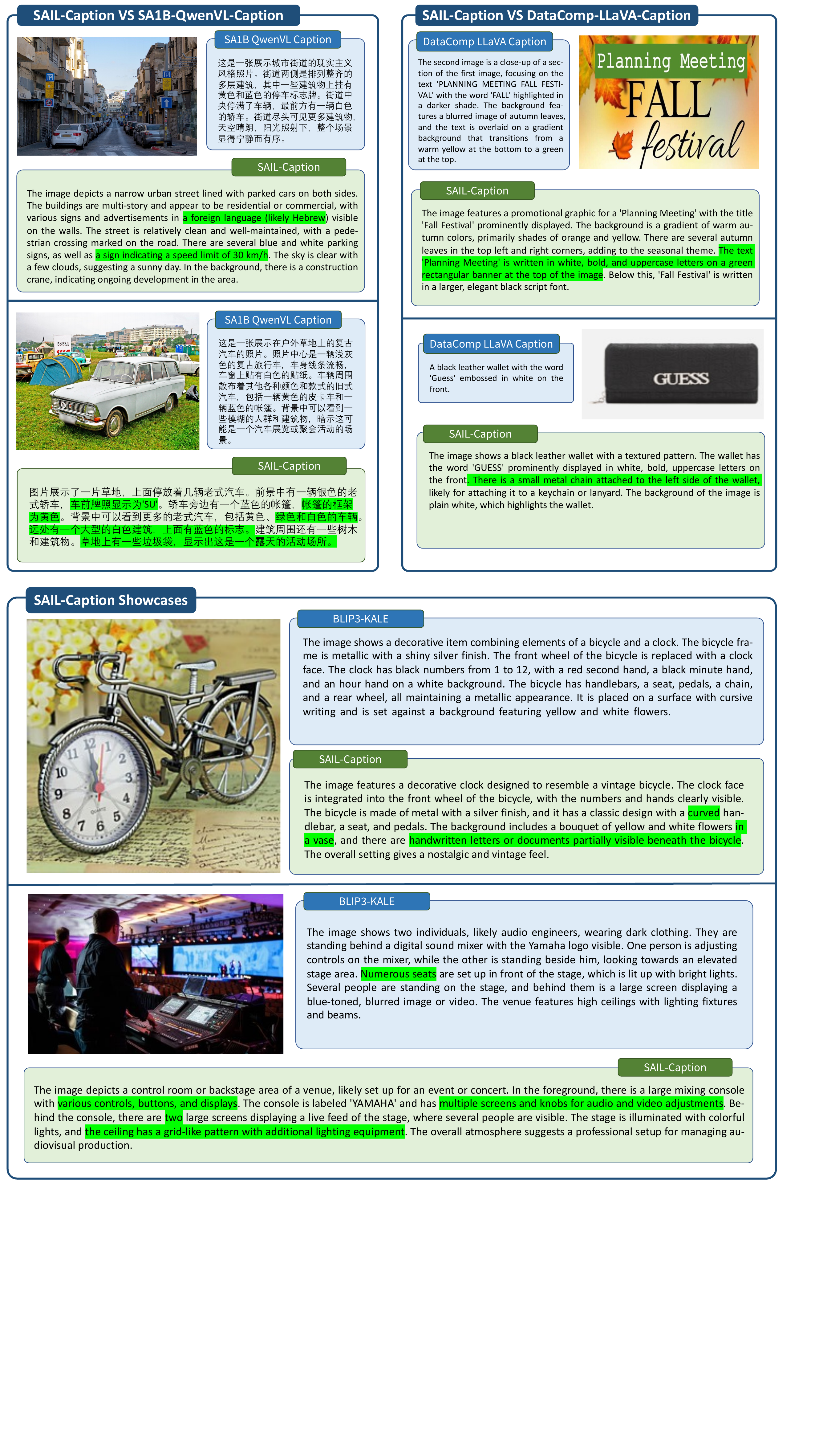}};
\end{tikzpicture}
}
\caption{
SAIL-Caption showcases versus SA1B-QwenVL-Caption, DataComp-LLaVA-Caption, and BLIP3-KALE. 
Images are curated from SA1B, DataComp and BLIP3-KALE.
}
\label{fig: sailcaption_showcase}
\end{figure*}

We curate several image samples from SA1B~\cite{kirillov2023segment}, DataComp~\cite{gadre2024datacomp}, and BLIP3-KALE~\cite{awadalla2024blip3} as demonstrations to compare the quality of SAIL-Caption with existing opensource caption datasets. 
We compare SAIL-Caption with SA1B-QwenVL-Caption, DataComp-LLaVA-Caption, and BLIP3-KALE.
Showcases are shown in Figure~\ref{fig: sailcaption_showcase}. 
As the demonstrations show, SAIL-Caption encompasses more detailed visual elements than other alternative datasets in both English and Chinese.
Observations drawn from these showcases coincide with quantified caption quality evaluation results shown in Section~\ref{sec: appendix_sailcaption_quality}, underscoring the leading data quality of SAIL-Caption.

A subset of the SAIL-Caption dataset with considerable data size will be released to promote opensource VLM research.

\section{Visual Understanding Benchmark}
\label{sec: appedix_understanding_bmk}
To inspect SAIL-VL's visual understanding performance during pretraining stages, we curate a series of visual understanding benchmarks for evaluation. 
To be specific, we focus on evaluating model performance in detailed captioning and OCR tasks in both English and Chinese, which are also the optimization objectives of SAIL-VL and opensource VLMs' pretraining stages. 
We list the basic information of the selected visual understanding benchmarks in Table~\ref{tbl: understanding_benchmark}.

\paragraph{Benchmarks.}
DetailCaps-4870~\cite{dong2024benchmarking} encompasses images from a wide range of publicly available datasets, including COYO~\cite{kakaobrain2022coyo-700m}, LAION~\cite{schuhmann2021laion}, CC~\cite{changpinyo2021conceptual}, Flickr~\cite{young2014image}, SBU~\cite{ordonez2011im2text}, and COCO~\cite{chen2015microsoft}, as well as ground truth detail captions generated by powerful VLM APIs.
We use the human-refined version of DetailCaps-4870 for evaluation and adopt both the corrected Chinese captions and translated English captions for multilingual evaluation. 

\begin{table}[t]
\resizebox{0.45\textwidth}{15mm}{
\begin{tabular}{l c c c}
\toprule
Benchmark & Task type & Language & \# Sample \\
\midrule

\textit{Caption} \\
\quad DetailCaps-4870-EN~\cite{dong2024benchmarking} & Caption & EN & 4870 \\
\quad DetailCaps-4870-CN~\cite{dong2024benchmarking} & Caption & CN & 4870 \\

\textit{OCR} \\
\quad IDL-WDS~\cite{biten2022ocr} & OCR & EN & 1000 \\
\quad DocStruct~\cite{wang2020docstruct} & OCR & EN & 1000 \\
\quad SynthText~\cite{Gupta16} & OCR & EN & 1000 \\
\quad SynthDog-EN~\cite{kim2022donut} & OCR & EN & 1000 \\
\quad SynthDog-ZH~\cite{kim2022donut} & OCR & CN & 1000 \\

\bottomrule
\end{tabular}
}
\centering
\caption{
Basic information of the visual understanding benchmarks used in our experiments. 
}
\label{tbl: understanding_benchmark}
\end{table}

The remaining OCR benchmarks consist of images with a diverse distribution. 
IDL-WDS~\cite{biten2022ocr} consists of document pages with abundant text information; 
DocStruct~\cite{wang2020docstruct} contains both document pages but also illustrative images rendered from tables and charts; 
SynthText~\cite{Gupta16} is composed of images with a single word, but the fonts vary from one sample to another; 
SynthDog-EN and SynthDog-ZH~\cite{kim2022donut} are compositional datasets comprised of natural image backgrounds and foreground word pieces.

\paragraph{Metrics.}
We evaluate SAIL-VL's caption performance on the DetailCaps-4870 benchmark with GPT evaluation. 
Provided with three ground truth captions and a candidate caption, GPT is tasked to score the candidate caption based on the precision and recall of the visual elements. 
For OCR tasks, we compute the ANLS score~\cite{biten2019scene} between the predicted OCR contents and the ground truth ones, resulting in scores ranging from 0 to 1.
The higher score indicates better prediction quality. 

All benchmark data is curated from left-out subsets to avoid data leakage between model training and evaluation. 
We select a 500-case subset randomly from each benchmark to evaluate the pretrained model checkpoints for efficiency.

\section{SFT Data Quality Evaluation}
\label{sec: appendix_sft_data_eval}

\begin{figure*}[h!]
\centering
\resizebox{\textwidth}{!}{
\begin{tikzpicture}
\draw (0,0) node[inner sep=0] {\includegraphics[width=\columnwidth, trim={0.25cm 26cm 0.25cm 0cm}, clip]{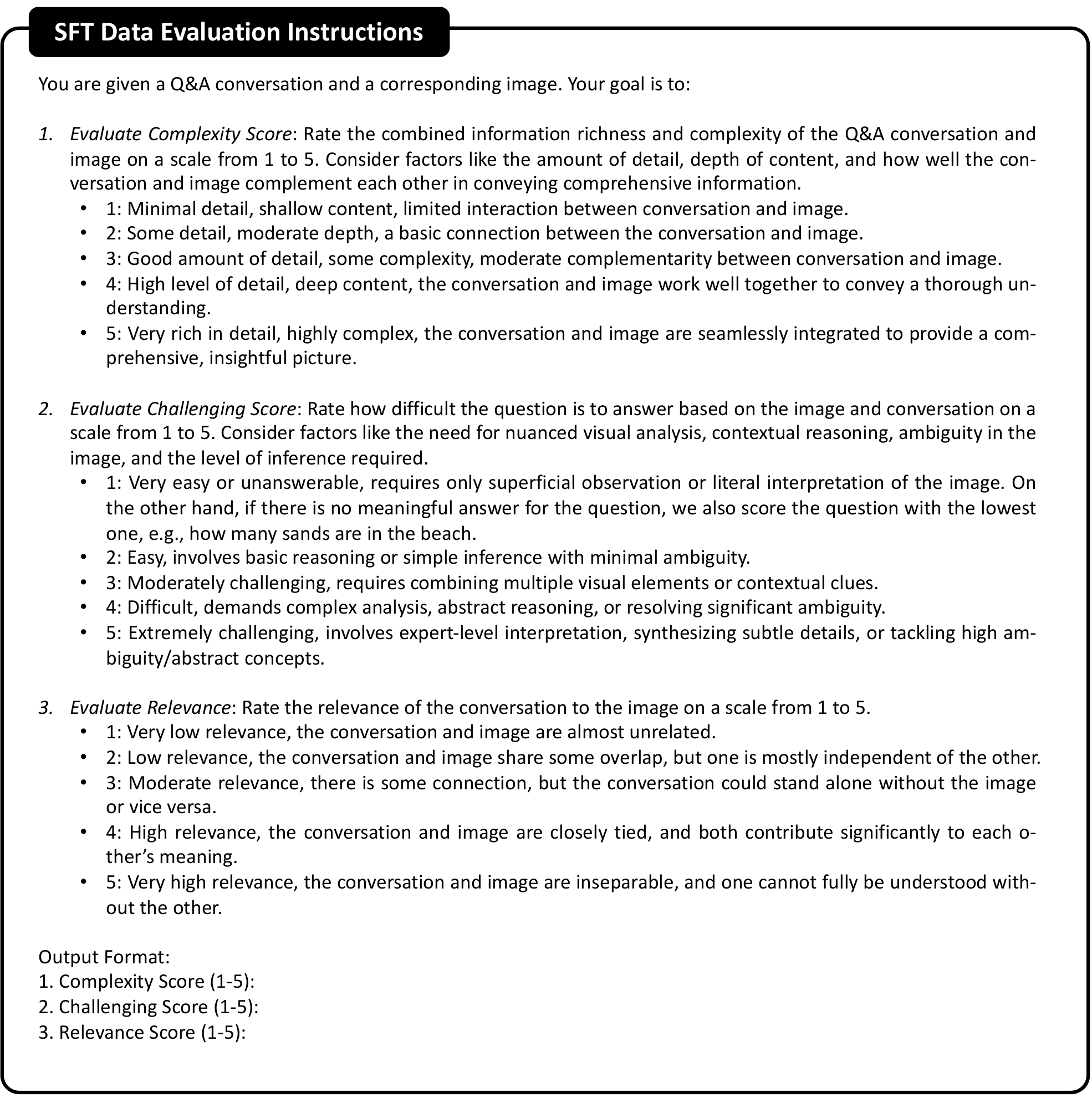}};
\end{tikzpicture}
}
\caption{
Detailed instructions for human experts to judge SFT data quality.
}
\label{fig: sft_eval_prompt}
\end{figure*}

In this section, we show the detailed instructions for SFT data quality evaluation.
As shown in Figure~\ref{fig: sft_eval_prompt}, human experts to annotate the challenging, complexity, and relevance scores for our three-stage SFT data. 

\section{Generalizing Visual Understanding Abilities to Instruction Following Tasks}
\label{sec: appendix_ptsft_correlation}

In this part, we investigate the correlation between VLM's visual understanding and instruction following abilities. 
We train SAIL-VL pretrained checkpoints from the pretrain-advance stage with exponentially larger training data sizes, and train them through either LLaVA-Next~\cite{liu2024llavanext} SFT data or our SFT-instruction data.

\begin{figure*}[t]
\centering
\resizebox{\textwidth}{53mm}{
\begin{tikzpicture}
\draw (0,0) node[inner sep=0] {\includegraphics[width=\columnwidth, trim={0cm 0cm 0cm 0cm}, clip]{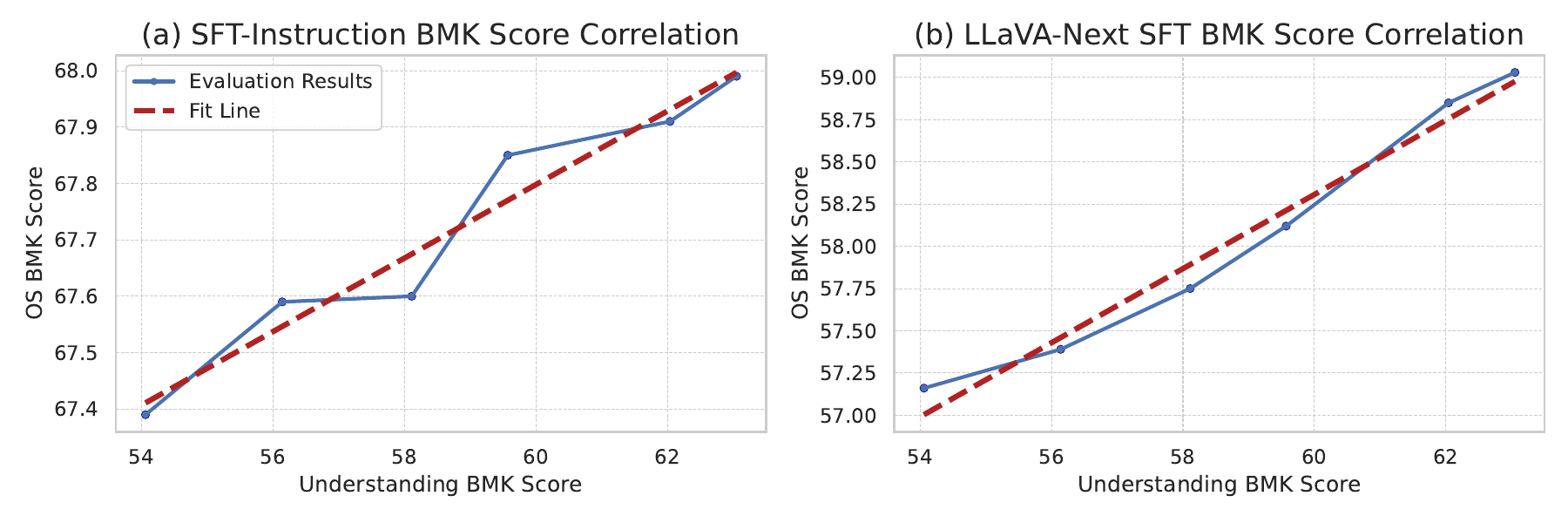}};
\end{tikzpicture}
}
\caption{
The correlation between SAIL-VL pretrained checkpoints' understanding performance and their performance on opensource benchmarks after SFT. 
``OS BMK Score” stands for average score on opensource benchmarks in our evaluations. 
}
\label{fig: understanding_correlation}
\end{figure*}

We plot the correlation of pretrained models' visual understanding performance and SFT models' opensource benchmark performance in Figure~\ref{fig: understanding_correlation}. 
A notable correlation is observed across different training strategies. 
As the VLM gains stronger visual understanding abilities during pretraining, its visual instruction following performance after SFT is improved accordingly, even if trained with different visual instruction tuning datasets. 
We quantify this correlation with Pearson correlation ($\rho$) and coefficient of determination ($R^2$). 
It turns out that SAIL-VL's pretrained visual understanding performance and SFT visual instruction following performance share a significant correlation. 
In experiments with SFT-Instruction data collection, pretrained model performance and SFT model scores share a Pearson correlation coefficient $\rho=0.97$ and a coefficient of determination $R^2=0.94$.
For LLaVA-Next SFT data experiments, we observe an even stronger correlation with $\rho=0.99$ and $R^2=0.98$.
These correlation results illustrate the generalization of model abilities across training stages and tasks, validating the necessity of pretraining VLMs for more robust visual understanding abilities. 

\section{Experiment Details}
\label{sec: appendix_exp_detail}

\subsection{Experiment settings}
\label{sec: appendix_exp_detail_setting}
\paragraph{Storage.} 
The training data of SAIL-VL's pretraining and SFT stages is stored on our Hadoop file system (HDFS) for persistent storage. 
Training data is fetched in a stream fashion during model training, making possible training with large scale distributed data storage. 

\paragraph{Training framework.}
We use PyTorch~\cite{paszke2019pytorch,ansel2024pytorch} version 2.1.0 with CUDA~\cite{cuda} 12.1 for model training. 
Deepspeed~\cite{rasley2020deepspeed} version 0.14.5 is used for SAIL-VL training. 
Flash-attention~\cite{dao2022flashattention,dao2023flashattention} implemented for 910B NPU~\cite{npu} is leveraged for fast attention computation. 

We process training data sequences with a stream accumulator, which packs sequences in a micro batch into a long sequence for model training. 
This strategy speeds up SAIL-VL model training by approximately $40\%$.


\paragraph{Training resources.} 
We conduct experiments with Huawei 910B x86 NPU~\cite{ascend}.
To train the SAIL-VL-2B model, we allocate 90,053 NPU hours for pretraining and 10,992 NPU hours for SFT stages, resulting in a total of 101,045 NPU hours in model training. 
For the SAIL-VL-8B model, we use 26M samples in pretraining for efficiency, and the same data collections as the 2B model are used in SFT stages. 
The 8B model consumes 19,575 NPU hours to train, where 6,672 and 12,903 NPU hours are allocated in pretraining and SFT stages, respectively. 

\subsection{Experiment Results. }
\label{sec: appendix_exp_detail_result}
We show model training details for both SAIL-VL-2B and SAIL-VL-8B models in Table~\ref{tbl: all_2b} and Table~\ref{tbl: all_8b}. 
We add InternVL2~\cite{internvl2}, InterVL2.5~\cite{chen2024expanding}, and Aquila~\cite{gu2024infinity} series models as supplementary baselines for evaluation.

\begin{table*}[t]
\resizebox{\textwidth}{!}{
\begin{tabular}{l | c c c c c c c}
\toprule

Benchmark & SAIL-VL & Qwen2-VL & InternVL2 & InternVL2.5 & InternVL2.5-MPO & Aquila & DeepSeekVL-2-Tiny \\

\midrule
\rowcolor{gray!15} \multicolumn{8}{c}{\textit{Overall Performance}} \\
\midrule
Opensource Average        & \textbf{69.1} & 64.4 & 62.9 & 66.5 & \underline{67.7} & 66.5 & 67.0  \\
\quad General VQA        & 60.4 & 58.3 & 55.6 & \underline{62.6} & \textbf{63.1} & 59.8 & 59.4  \\
\quad OCR VQA        &\textbf{75.9} &72.5 &68.2 &68.7 & 71.1 &71.7 &\underline{74.4}  \\
\quad Math\&Knowledge        & \textbf{79.0} & 59.0 & 70.6 & 73.0 & 75.3 & \underline{75.4} & 71.3  \\
\quad Hallucination        & \textbf{66.2} & 62.9 & 61.7 & \underline{66.2} & 64.5 & 62.9 & 63.6 \\

\midrule
\rowcolor{gray!15} \multicolumn{8}{c}{\textit{General VQA}} \\
\midrule
MMStar~\citeyearpar{chen2024we}        & \textbf{55.1} & 46.3 & 50.5 & 53.5 & 54.3 & \underline{54.7} & 49.9  \\
$\text{MMBench}_{\text{DEV}}$~\citeyearpar{liu2024mmbench}        & 72.4 & 68.8 & 70.3 & \underline{73.1} & 72.5 & \textbf{74.4} & 68.3  \\
$\text{MMMU}_{\text{VAL}}$~\citeyearpar{yue2024mmmu}        & 40.1 & 39.9 & 34.2 & 40.7 & \underline{41.2} & \textbf{44.1} & 39.6  \\
MME~\citeyearpar{fu2023mme}        & 1969 & 1923 & 1859 & \underline{2090} & \textbf{2123} & 1808 & 1910  \\
$\text{SEEDBench}_{\text{IMG}}$~\citeyearpar{li2023seed}        & \textbf{74.7} & 72.0 & 70.9 & 73.4 & 73.2 & \underline{73.9} & 72.5  \\
RealWorldQA~\citeyearpar{grok15_v}        & 63.8 & 60.9 & 56.7 & 60.9 & 60.7 & \underline{64.1} & \textbf{64.8}  \\
MMVet~\citeyearpar{yu2024mm}        & 46.1 & 51.2 & 40.4 & \underline{61.7} & \textbf{64.0} & 42.7 & 52.8  \\

\midrule
\rowcolor{gray!15} \multicolumn{8}{c}{\textit{OCR VQA}} \\
\midrule
$\text{AI2D}_{\text{TEST}}$~\citeyearpar{kembhavi2016diagram}        & \textbf{79.0} & 72.3 & 74.2 & 75.0 & \underline{75.3} & 75.0 & 74.6  \\
$\text{DocVQA}_{\text{VAL}}$~\citeyearpar{mathew2021docvqa}        & \textbf{89.2} & \underline{88.7} & 86.0 & 87.4 & 87.8 & 85.0 & 88.6  \\
$\text{InfoVQA}_{\text{VAL}}$~\citeyearpar{mathew2022infographicvqa}        & \textbf{67.2} & 63.4 & 57.5 & 61.6 & 61.6 & 60.5 & \underline{63.8}  \\
$\text{ChartQA}_{\text{TEST}}$~\citeyearpar{masry2022chartqa}        & \underline{81.0} & 70.6 & 71.7 & 73.3 & 70.9 & 76.6 & \textbf{81.2}  \\
$\text{TextVQA}_{\text{VAL}}$~\citeyearpar{singh2019towards}        & 75.7 & \underline{78.8} & 73.4 & 76.4 & 77.2 & 76.4 & \textbf{80.5}  \\
$\text{OCRVQA}_{\text{TEST}}$~\citeyearpar{mishraICDAR19}        & \textbf{58.5} & \underline{54.3} & 36.2 & 28.3 & 40.0 & 51.3 & 51.4  \\
OCRBench~\citeyearpar{liu2024ocrbench}        & 806 & 794 & 786 & 789 & \textbf{846} & 772 & \underline{808}  \\

\midrule
\rowcolor{gray!15} \multicolumn{8}{c}{\textit{Math\&Knowledge}} \\
\midrule
$\text{MathVista}_{\text{MINI}}$~\citeyearpar{lu2023mathvista}        & \textbf{62.8} & 45.0 & 46.8 & 51.1 & 55.3 & \underline{59.4} & 54.5  \\
$\text{ScienceQA}_{\text{VAL}}$~\citeyearpar{saikh2022scienceqa}        & \underline{95.3} & 73.0 & 94.4 & 94.9 & \underline{95.3} & 91.4 & 88.1  \\

\midrule
\rowcolor{gray!15} \multicolumn{8}{c}{\textit{Hallucination}} \\
\midrule
HallusionBench~\citeyearpar{guan2024hallusionbench}        & \textbf{45.7} & 38.3 & 38.2 & \underline{42.5} & 39.2 & 42.1 & 38.4  \\
POPE~\citeyearpar{Li-hallucination-2023}        & 86.7 & 87.6 & 85.3 & \textbf{89.9} & \underline{89.8} & 83.6 & 88.8  \\

\bottomrule
\end{tabular}
}
\centering
\vspace{0mm}
\caption{
Complete evaluation results for SAIL-VL-2B and opensource VLMs of comparable sizes. 
Denotations are defined the same as Table~\ref{tbl: main}.
}
\label{tbl: all_2b}
\end{table*}

\begin{table*}[t]
\resizebox{\textwidth}{65mm}{
\begin{tabular}{l | c c c c c c}
\toprule

Benchmark & SAIL-VL & Qwen2-VL & InternVL2 & InternVL2.5 & InternVL2.5-MPO & DeepSeekVL-2-Small \\

\midrule
\rowcolor{gray!15} \multicolumn{7}{c}{\textit{Overall Performance}} \\
\midrule

Opensource Average        & \textbf{74.5} & 73.0 & 70.0 & 73.2 & \underline{74.3} & 72.7  \\
\quad General VQA        & 68.3 & 68.5 & 66.6 & \underline{70.1} & \textbf{71.2} & 66.8  \\
\quad OCR VQA        & \textbf{79.8} & \underline{79.6} & 72.6 & 75.0 & 76.3 & 79.0 \\
\quad Mah\&Knowledge & \textbf{83.3} & 71.0 & 78.4 & 81.5 & \underline{83.2} & 79.0  \\
\quad Hallucination        & 68.7 & 67.5 & 64.5 & \underline{69.5} & \textbf{69.7} & 65.3 \\

\midrule
\rowcolor{gray!15} \multicolumn{7}{c}{\textit{General VQA}} \\
\midrule
MMStar~\citeyearpar{chen2024we}        & \underline{64.2} & 58.3 & 61.6 & 62.5 & \textbf{65.3} & 57.7  \\
$\text{MMBench}_{\text{DEV}}$~\citeyearpar{liu2024mmbench}        & 79.5 & 79.5 & 80.3 & \underline{83.1} & \textbf{83.3} & 78.1  \\
$\text{MMMU}_{\text{VAL}}$~\citeyearpar{yue2024mmmu}        & 48.2 & 50.9 & 47.6 & \underline{52.4} & \textbf{52.8} & 47.6  \\
MME~\citeyearpar{fu2023mme}        & 2244 & 2321 & 2215 & \textbf{2339} & 2321 & 2149  \\
$\text{SEEDBench}_{\text{IMG}}$~\citeyearpar{li2023seed}        & 75.5 & 75.3 & 75.4 & \textbf{77.0} & \underline{76.9} & 76.8  \\
RealWorldQA~\citeyearpar{grok15_v}        & \textbf{71.9} & 69.7 & 64.7 & 69.9 & 70.2 & 70.2  \\
MMVet~\citeyearpar{yu2024mm}        & 58.3 & \underline{62.6} & 57.7 & 62.1 & \textbf{66.8} & 60.3  \\

\midrule
\rowcolor{gray!15} \multicolumn{7}{c}{\textit{OCR VQA}} \\
\midrule
$\text{AI2D}_{\text{TEST}}$~\citeyearpar{kembhavi2016diagram}        & 83.7 & 82.9 & 83.7 & \textbf{84.6} & \underline{84.1} & 82.0  \\
$\text{DocVQA}_{\text{VAL}}$~\citeyearpar{mathew2021docvqa}        & 92.2 & \textbf{93.7} & 90.8 & 91.8 & 92.1 & \underline{92.3}  \\
$\text{InfoVQA}_{\text{VAL}}$~\citeyearpar{mathew2022infographicvqa}        & 75.2 & \underline{75.9} & 61.5 & 75.5 & \textbf{76.2} & 72.5  \\
$\text{ChartQA}_{\text{TEST}}$~\citeyearpar{masry2022chartqa}        & \underline{84.6} & 81.6 & 82.0 & 82.9 & 77.6 & \underline{84.6}  \\
$\text{TextVQA}_{\text{VAL}}$~\citeyearpar{singh2019towards}        & 77.7 & \textbf{83.8} & 77.6 & 79.0 & 79.2 & \underline{83.3}  \\
$\text{OCRVQA}_{\text{TEST}}$~\citeyearpar{mishraICDAR19}        & \textbf{61.4} & \underline{56.2} & 38.1 & 29.5 & 36.7 & 54.5  \\
OCRBench~\citeyearpar{liu2024ocrbench}        & \underline{835} & 833 & 746 & 819 & \textbf{880} & 834  \\

\midrule
\rowcolor{gray!15} \multicolumn{7}{c}{\textit{Math\&Knowledge}} \\
\midrule
$\text{MathVista}_{\text{MINI}}$~\citeyearpar{lu2023mathvista}        & \underline{68.4} & 57.3 & 59.4 & 65.4 & \textbf{68.5} & 61.8  \\
$\text{ScienceQA}_{\text{VAL}}$~\citeyearpar{saikh2022scienceqa}        & \textbf{98.2} & 84.6 & 97.4 & 97.6 & \underline{97.9} & 96.2  \\

\midrule
\rowcolor{gray!15} \multicolumn{7}{c}{\textit{Hallucination}} \\
\midrule
HallusionBench~\citeyearpar{guan2024hallusionbench}        & \textbf{52.2} & 48.5 & 44.6 & 50.1 & \underline{50.3} & 41.2  \\
POPE~\citeyearpar{Li-hallucination-2023}        & 85.2 & 86.5 & 84.4 & 88.8 & \underline{89.1} & \textbf{89.4}  \\

\bottomrule
\end{tabular}
}
\centering
\vspace{0mm}
\caption{
Complete evaluation results for SAIL-VL-8B and opensource VLMs of comparable sizes. 
Denotations are defined the same as Table~\ref{tbl: main}.
}
\label{tbl: all_8b}
\end{table*}

\end{document}